\documentclass[Afour,sageh,times]{sagej}

\usepackage{moreverb,url}
\usepackage{bm}
\usepackage{mathrsfs}
\usepackage{graphicx}
\usepackage{subcaption}
\usepackage{rotating}
\usepackage{amsmath}
\usepackage{multirow}
\usepackage{multicol}
\usepackage[colorlinks,bookmarksopen,bookmarksnumbered,citecolor=red,urlcolor=red]{hyperref}
\usepackage{natbib}

\newcommand\BibTeX{{\rmfamily B\kern-.05em \textsc{i\kern-.025em b}\kern-.08em
T\kern-.1667em\lower.7ex\hbox{E}\kern-.125emX}}

\def\volumeyear{2016}

\begin{document}

\runninghead{Giuseppe Loffredo et al}

\title{Enhancing Visual Interpretability and Explainability in Functional Survival Trees and Forests}

\author{Giuseppe Loffredo*\affilnum{1}, Elvira Romano\affilnum{1} and Fabrizio Maturo\affilnum{2}}

\affiliation{
\affilnum{1}Department of Mathematics and Physics, University of Campania Luigi Vanvitelli, Caserta, Italy\\
\affilnum{2}Department of Economics, Statistics and Business, Universitas Mercatorum, Rome, Italy
}

\begin{abstract}
Functional survival models are key tools for analyzing time-to-event data with complex predictors, such as functional or high-dimensional inputs. Despite their predictive strength, these models often lack interpretability, which limits their value in practical decision-making and risk analysis.
This study investigates two key survival models: the Functional Survival Tree (FST) and the Functional Random Survival Forest (FRSF). It introduces novel methods and tools to enhance the interpretability of FST models and improve the explainability of FRSF ensembles. Using both real and simulated datasets, the results demonstrate that the proposed approaches yield efficient, easy-to-understand decision trees that accurately capture the underlying decision-making processes of the model ensemble.
\end{abstract}

\keywords{Functional Data Analysis, Survival Analysis, Functional Survival Tree, Functional Random Survival Forests, Functional Survival Discrimination Curve.}
\maketitle

\section*{List of Acronyms}

\begin{tabular}{@{}lp{0.72\linewidth}@{}}
\textbf{CHF}        & Cumulative Hazard Function \\
\textbf{FCT}        & Functional Classification Tree \\
\textbf{FDA}        & Functional Data Analysis \\
\textbf{FPC}        & Functional Principal Component \\
\textbf{FPCA}       & Functional Principal Component Analysis \\
\textbf{FPSP}       & Functional Predicted Separation Prototype \\
\textbf{FSDC}       & Functional Survival Discrimination Curve \\
\textbf{FRSF}       & Functional Random Survival Forest \\
\textbf{FST}        & Functional Survival Tree \\
\textbf{IB}         & In-Bag (training sample in random forests) \\
\textbf{LFSDC}      & Local Functional Survival Discrimination Curve \\
\textbf{MTGD}       & Mean Time Global Difference \\
\textbf{MTNGD}      & Mean Time Normalized Global Difference \\
\textbf{OOB}        & Out-of-Bag (validation sample in random forests) \\
\textbf{PACE}       & Principal Analysis by Conditional Expectation \\
\textbf{PFI}        & Permutation Feature Importance \\
\textbf{RSF}        & Random Survival Forest \\
\textbf{SF}         & Survival Function \\
\textbf{SHAP}       & Shapley Additive Explanations \\
\textbf{SOFA}       & Sequential Organ Failure Assessment \\
\textbf{ST}         & Survival Tree \\
\textbf{TSD}        & Time SHAP Difference \\
\textbf{TNSD}       & Time Normalized SHAP Difference \\
\textbf{SurvSHAP(t)} & Time-dependent Shapley Additive Explanation in Survival Models \\
\textbf{VIMP}       & Variable Importance (in Random Forests) \\
\end{tabular}

\section{Introduction}
Tree-based models are fundamental in machine learning, providing powerful tools for both classification and regression. These models iteratively partition data based on feature-specific cutoff values, creating a hierarchical structure of internal and terminal (leaves) nodes. The final prediction is typically obtained by aggregating the outcomes within each leaf, using the mean for regression tasks and majority voting for classification \citep{Hastie2001}.

With technological advancements, high-dimensional temporal data have become increasingly common, requiring more sophisticated analytical frameworks. Functional Data Analysis (FDA) \citep{Ramsay2005} offers a natural approach to modelling such data, treating entire curves or trajectories as single entities rather than discrete points. This paradigm has driven the evolution and extension of tree-based methods to accommodate functional inputs, leading to new challenges in interpretability and decision-making.

\citet{Balakrishnan2006} developed an intuitive decision tree model for functional inputs, demonstrating its accuracy and interpretability. Moreover, \citet{Nerini2007} explored functional regression trees to predict probability density functions, while \citet{Brandi2018} introduced a method using tree-based classification for functional data, relying on a distance correlation function. In addition, \citet{Belli2021} utilised constrained convex optimisation to learn a weighted functional space, extracting multiple integral characteristics for defining splits in binary trees. Recently, \citet{GOLOVKINE2022107376} proposed a functional tree model for clustering multiple functional variables in an unsupervised manner. Meanwhile, \citet{MaturoVerde} integrated random forests with binary trees to classify biomedical signals in a supervised approach, \citet{Hael2023} proposed an unbiased recursive decision tree methodology, extending the binary decision tree framework for supervised learning tasks involving functional data, particularly in the context of electrocardiogram signal classification.

In recent years, Random Survival Forests (RSF) have attracted growing attention in the medical domain due to their flexibility in modelling time-to-event data beyond the constraints of traditional Cox models. Extensions of RSF have addressed increasingly complex settings: accommodating time-varying covariates \citep{Yao20222217}, dealing with dependent censoring through copula-based estimators and new aggregation rules \citep{Moradian2019445}, and incorporating high-dimensional longitudinal predictors within competing-risk frameworks \citep{Devaux20232331}.

In the context of Functional Data Analysis (FDA), Functional Random Survival Forests (FRSF) \citep{Lin202199, Loffredo2025} represent a natural and innovative extension of RSF for survival data characterized by functional or longitudinal predictors. For example, \citet{Lin202199} proposes a dynamic prediction framework that integrates multivariate longitudinal trajectories via functional principal component analysis and sparse covariance estimation, significantly improving predictive performance in complex clinical scenarios such as Alzheimer’s disease.
More recently, \citet{Loffredo2025} introduced an FRSF model based on a new data structure termed Censored Functional Data (CFD), specifically designed to handle time-to-event data characterized by both censoring and irregular temporal patterns. This approach enables a more accurate representation of functional survival trajectories and enhances prediction across heterogeneous subgroups.

To the best of our knowledge, although interpretability tools from traditional decision trees have been explored in classical RSF settings \citep{deng2019interpreting}, they have not yet been formally adapted to the functional domain. In this direction, Functional Survival Trees (FSTs) offer a promising and intuitive framework for visualising and understanding node-level survival dynamics when functional covariates are involved. 
Nonetheless, FST and FRSF face limitations in interpretability and explainability, particularly in understanding how functional features contribute to model decisions over time.
This work addresses these challenges by proposing novel methods to improve the transparency of tree-based survival models for functional data, positioning our study within the broader scope of explainable artificial intelligence (XAI) in the medical domain.
In the context of FSTs, this research focuses on enhancing interpretability by introducing tools such as the Local Survival Discrimination Curve (LSDC), as separability metrics and visual representation of survival curves at terminal nodes. These instruments clarify the role of specific features in survival predictions and illustrate how tree splits affect survival outcomes. For FRSF models, we shift the emphasis from interpretability to explainability, considering the complexity of ensemble methods. Rather than analyzing individual trees, we investigate how different features collectively influence predictions across the functional forest.

The paper is organized as follows. Section 2 reviews the state of the art on Functional Survival Trees (FSTs) and Functional Random Survival Forests (FRSFs). Section 3 presents our methodological contributions aimed at improving interpretability and explainability in these models. Section 4 reports the numerical results, including simulations and a real-world case study based on the SOFA dataset.

\section{Material and Methods}

\subsection{Functional Tree Models in Survival Analysis}

Functional tree models are widely used in survival analysis to handle time-to-event data collected during follow-up studies denoted as $\mathcal{T}$. These models leverage functional data to capture temporal patterns, enhancing predictive accuracy. However, their effectiveness depends on the ability to interpret decision-making processes and explain the role of functional components within the model. The FSTs and FRSF extend the classical Survival Trees (STs) and Random Survival Forest (RSF) by using the Functional Principal Component scores (FPCs) \citep{Ramsay2005}, derived through the Functional Principal Component Analysis (FPCA), as predictors. Unlike classical Functional Classification Trees (FCTs) \citep{maturo2023supervised}, which aim to predict categorical outcomes, Functional Survival Trees (FSTs) are designed to model survival probabilities over time and can handle irregular data where not all subjects experience the event within the study period \citep{Loffredo2025}.
 
In a survival study, denoted as $\mathcal{T}$, the time-dependent data  $\bm{Y}_i = (Y_{i1}, \ldots, Y_{iJ_i})^T$ is recorded for the $i$-th subject,  where the observations $Y_i=Y(t_i)$ is evaluated in $t_i\in\mathcal{T}$ until the survival time $T_i^* = \min(C_i, T_i)$ is reached. The $C_i$ and $T_i$ denote the censoring time and the event time, respectively. Notably, the number of observations $J_i$ may differ across $N$ subjects, such that $J_i\neq J_k$ for some $i\neq k$. The method starts considering the observed survival data $(Z_i, \bm{Y}_i, T_i^*, \delta_i)$ collected for $i=1,\ldots,N$ to construct the model prediction, where $Z_i$ denotes the binary response and $\delta_i$ represents the censoring indicator, given by $\delta_i=\mathbf{1}_{[T_i^*=T_i]}$. 
 
The construction of FST considers the pairs $\{\bm{Z}, \mathcal{X}\}$ for modelling.  Here, $\bm{Z} \in \{0,1\}^N$ indicates whether the event occurred ($Z_i = 1$) or not occurred ($Z_i = 0$) for the $i$-th subject, while $\mathcal{X} = \{X_i(t): t \in \mathcal{T}, \, i = 1, \ldots, N\}$ denotes the observed functional variable obtained from the irregular time data $\bm{Y}_i\in\mathbb{R}^{J_i}$.
 To handle the data's irregularity, the method uses the Functional Principal Component Analysis by Conditional Expectation (PACE) decomposition \citep{yao2005} to extract meaningful information from the observations, transforming them into a functional form.  
 
The method reconstructs functional data non-parametrically, defining the mean function $\mu(t)$ and covariance function $G(s, t) = \sum_{m=1}^\infty \lambda_m \xi_m(t) \xi_m(s)\quad\text{with}\quad t,s\in\mathcal{T}$, which undergo spectral decomposition. The mean function $\mu(t)$ and covariance function $G(s, t)$ are estimated using local linear smoothers, while eigenfunctions $\xi_m(t)$ and eigenvalues $\lambda_m$ are derived from the smoothed covariance surface. The functional data $X(t)$ can be obtained in terms of principal components starting from the Karhunen-Loève expansion. Thus, given irregularly observed data $(Y_{ij}, t_{ij})$ recorded for $N$ subjects and considering the measurement errors $\varepsilon_{ij}$, the expansion expresses $X(t)$ as follows:   

\begin{align}
    &Y_{ij} = X_i(t_{ij}) + \varepsilon_{ij} 
    = \mu(t_{ij}) + \sum_{m=1}^{p} \nu_{im} \xi_m(t_{ij}) + \varepsilon_{ij}\notag\\
    &\text{with} \quad t_{ij} \in \mathcal{T} 
\end{align}
where $\{\nu_{im}\}_{m=1}^p$ is a set of Functional principal component scores (FPCs), while $\{\xi_{m}\}_{m=1}^p$ denotes the associated eigenfunctions, which are derived according to the principles of orthonormality.
 The FPCs' scores \(\nu_{im}\) are computed under joint Gaussianity assumptions obtaining:

\begin{equation}
\hat{\nu}_{im} = \hat{\lambda}_m \hat{\bm{\xi}}_m^T \hat{\bm{\Sigma}}_{\bm{Y}_i}^{-1} (\bm{Y}_i - \hat{\bm{\mu}}_i),\label{eq:score}
\end{equation}

\noindent where $\bm{Y}_i$ is the observed data vector for subject $i$, $\hat{\bm{\Sigma}}_{\bm{Y}_i}$ is the covariance matrix, and $\hat{\bm{\xi}}_m$ is the eigenfunction vector evaluated at observed times. Finally, the predicted trajectory $\hat{X}_i(t)$ for the $i$-th unit is given by:

\begin{equation}
\hat{X}_i(t) = \hat{\mu}(t) + \sum_{m=1}^p \hat{\nu}_{im} \hat{\xi}_m(t) \quad\text{with}\quad t \in \mathcal{T}\label{eq:predicted_trajectory}
\end{equation}
which represents the estimated trend over time $\mathcal{T}$, computed for each subject $i = 1, \ldots, N$.
These functional representations are then integrated into the functional tree-based models, serving as predictors for constructing FSTs and FRSF.

 To manage the functional data, the Functional Survival Tree with Principal Components (FSTs-FPCs) introduced by \citet{Loffredo2025} extends traditional ST by using the FPCs' scores as new features \citep{MaturoVerde, riccio2024supervised}. Thus, the ST is built considering the survival dataset  $D_N = \{Z_i, \bm{\hat{V}}_i, T_i^*, \delta_i\}_{i=1}^N$, where $\bm{\hat{V}}_i=(\hat{\nu}_{i1},\hat{\nu}_{i2},\ldots,\hat{\nu}_{ip})$ is the $i$-th score feature vector obtained from the Equation \ref{eq:score}. The score vector $\bm{\hat{V}}_i$ corresponds to the $i$-th row of the feature matrix, defined as follows:

\begin{equation}
\label{matrix_feature}
\bm{\hat{V}} =
\begin{pmatrix}
\hat{\nu}_{11} & \hat{\nu}_{12} & \dots & \hat{\nu}_{1p} \\
\hat{\nu}_{21} & \hat{\nu}_{22} & \dots & \hat{\nu}_{2p} \\
\vdots & \vdots & \ddots & \vdots \\
\hat{\nu}_{N1}& \hat{\nu}_{N2} & \dots & \hat{\nu}_{Np}
\end{pmatrix},
\end{equation}

\noindent with $\bm{\hat{V}} \in \mathbb{R}^{N \times p}$. In FCTs \citep{MaturoVerde}, the single tree growth uses splitting criteria that minimise impurity measures, such as the Gini index. FSTs adapt this to survival analysis, using survival-specific splitting criteria to maximise differences in survival curves \citep{Loffredo2025}. The tree construction partitions the $p$-dimensional feature space into non-overlapping terminal regions $\mathbb{A} = \{\mathcal{A}_s\}_{s \in \mathcal{S}}$, where $\mathcal{S}$ denotes the set of terminal nodes. These regions are mutually exclusive, i.e., $\mathcal{A}_s \cap \mathcal{A}_t = \emptyset$ for all $s \ne t$. The splits are determined using a survival-specific criterion, such as the log-rank test \citep{ziegler2007}, which aims to maximize survival differences between partitions by evaluating:

\begin{equation}
|L(\bm{\hat{V}}, c)| = 
\frac{\sum_{l=1}^L \left( d_{l1} - r_{l1} \frac{d_l}{r_l} \right)}
{\sqrt{\sum_{l=1}^L \frac{r_{l1}}{r_l} \left(1 - \frac{r_{l1}}{r_l} \right) 
\left(\frac{r_l - d_l}{r_l - 1} \right) d_l}},
\label{logrank}
\end{equation}

\noindent where $d_{l1}$ and $d_{l2}$ are the number of events in the two daughter nodes at time $t_l$, with $d_l = d_{l1} + d_{l2}$ as the total number of events. Similarly, $r_{l1}$ and $r_{l2}$ are the numbers at risk, with $r_l = r_{l1} + r_{l2}$. The optimal threshold $c^*$ and variable $\bm{\hat{V}}^*$ are selected by maximizing the value $|L(\bm{\hat{V}}^*, c^*)|$, such that $|L(\bm{\hat{V}}^*, c^*)| \geq |L(\bm{\hat{V}}, c)|$ for all $\bm{\hat{V}} \in \mathbb{R}^N$ and $c \in \mathbb{R}$.

The survival predictor in the terminal node $\mathcal{A}_s$ is estimated using the Cumulative Hazard Function (CHF) by the Nelson-Aalen estimator and the Survival Function (SF) by the Kaplan-Meier estimator given by:
\begin{equation}  
\hat{H}_{\mathcal{A}_s}(t) = \sum_{t_{ls} \leq t} \frac{d_{ls}}{r_{ls}} , \quad \hat{S}_{\mathcal{A}_s}(t) = \prod_{t_{ls} \leq t} \Bigg(1 -\frac{d_{ls}}{r_{ls}}\Bigg)\,
\label{estimators}
\end{equation}

\noindent where $d_{ls}$ and $r_{ls}$ are the number of events and individuals at risk at time $t_{ls}$ in $\mathcal{A}_s$.

To improve performance and reduce the variability of individual FSTs, the FRSF approach is introduced. FRSF constructs an ensemble of FSTs by aggregating multiple FSTs built on different subsets of the data and predictors. This ensemble approach reduces variance, enhances robustness, and improves prediction accuracy, particularly for complex survival data. By incorporating the FPCs' scores, FRSF effectively captures the structure of functional predictors, offering a more stable and reliable alternative to a single FST. The bootstrap samples, i.e, the In-Bag data (IB) are used for training, while out-of-bag (OOB) data are used to validate the model. The FSTs are grown by randomly selecting predictors and applying a splitting rule. Let $\mathbb{A}_{(b)}$ denote the set of partitions generated by the $b$-th tree, where $b = 1, \ldots, B$ corresponds to the tree built from the $b$-th bootstrap sample of the IB data. The ensemble's CHF and SF are computed by averaging the estimates from all individual trees as follows:
\begin{align}
&\overline{H}_{\{\mathbb{A}_{(b)}\}_{b=1}^B}^{IB}(t|\hat{\bm{V}}) = 
\frac{1}{B} \sum_{b=1}^B \hat{H}_{\mathbb{A}_{(b)}}^{IB}(t|\hat{\bm{V}})\notag\\
&\overline{S}_{\{\mathbb{A}_{(b)}\}_{b=1}^B}^{IB}(t|\hat{\bm{V}}) = 
\frac{1}{B} \sum_{b=1}^B \hat{S}_{\mathbb{A}_{(b)}}^{IB}(t|\hat{\bm{V}}),
\end{align}

\noindent where $\hat{H}_{\mathbb{A}_{(b)}}^{IB}(t|\hat{\bm{V}})=\sum_{s\in\mathcal{S}_b}\mathbf{1}(\hat{\bm{V}}\in\mathcal{A}_s^{b})\,\hat{H}_{\mathcal{A}_s^{b}}^{IB}(t)$ and 
$\hat{S}_{\mathbb{A}_{(b)}}^{IB}(t|\hat{\bm{V}})=\sum_{s\in\mathcal{S}_b}\mathbf{1}(\hat{\bm{V}}\in\mathcal{A}_s^{b})\,\hat{S}_{\mathcal{A}_s^{b}}^{IB}(t)$ are the CHF and SF calculated by Equation \ref{estimators} for individual FSTs.

Combining FDA and survival analysis within the FRSF framework offers an advanced way to handle irregularly spaced time data and censored observations. This integration improves predictive accuracy and creates a more interpretable model. 

While FSTs pose challenges in terms of interpretability at the level of individual trees, FRSF add further complexity regarding explainability, especially concerning how both FPC scores and traditional features, such as age or height, influence survival predictions across the ensemble. In particular, there is still considerable complexity in understanding the decision rules at each node, the application of a single survival tree to a study population, and the combined impact of FPCs' scores and classical features on the tree's construction. Additionally, interpreting the role of these mixed features in shaping decision boundaries and their effect on survival outcomes introduces further challenges. Hence, enhancing the interpretability and explainability of these methods represents a critical objective of this study and a key direction for future research.

\section{Our contribution}

We propose a twofold contribution to enhance model transparency: graphical methods to improve the interpretability of Functional Survival Trees (FSTs), and explainability tools to clarify feature contributions in Functional Random Survival Forests (FRSFs).

\subsection{Graphical interpretability for Functional Survival Tree (FST)}

In tree-based methodologies, a significant challenge frequently arises from model complexity, which can hinder graphical interpretability and complicate the understanding of underlying processes. This challenge is particularly significant when working with mixed data, where continuous and functional features are available. Under these conditions, interpreting individual trees becomes increasingly challenging, as complex and non-linear interactions among diverse features may arise. This complicates the extraction of meaningful insights. However, this issue can be overcome using advanced mathematical and graphical tools that provide a clearer and more intuitive interpretation of the results.
A classic ST recursively splits the units at each node based on covariates to maximize the difference in survival distributions between the resulting groups. The tree uses statistical tests, like the Log-rank test, for splitting decisions. Each terminal node of the tree represents a subgroup with its survival curve, showing how the event of interest is distributed over time.

Let $D_N=\{(Y_i,\bm{X}_i,\bm{\hat{V}}_i,\bm{T}_{i},\delta_i)\,:\, i=1,\ldots,N\}$ be a mixed survival dataset, where $\bm{X}_i\in\mathbb{R}^q$ is the feature-vector represented as $\bm{X}_i=(x_{i1},x_{i2},\ldots,x_{iq})$, while $\bm{\hat{V}}_i\in\mathbb{R}^p$ is the FPC score-vector generated by PACE decomposition and $\bm{T}_{i}\in\mathbb{R}^{J_i}$ is the time-vector for the $i$-th unit.
We define as $\mathrm{T}$ the Functional Mixed Survival Tree (FMST) generated by $D_N$ using the classical approach for building the FST. 

Our approach introduces key innovations that enhance the interpretability and robustness of the model in the interpretation of the time-dependent effects that may not be immediately apparent from numerical outputs into the functional space $\mathcal{L}^2(\mathcal{T})$. Specifically, we introduce the Local Functional Survival Discrimination Curve (LFSDC) to capture survival dynamics across different levels and within each node of the functional survival tree. This tool provides a graphical interpretation of distinct survival states and introduces a separability metric designed to quantify the effectiveness of each split. The metric highlights transitions between different stages of survival, enabling a clearer understanding of how the functional features contribute to survival differentiation along the tree structure.
The underlying idea of the LFSDC has been inspired by the definition of Functional Predicted Separation Prototype (FPSP) introduced by \citep{MaturoVerde}. 

Initially, starting from its generalization, which considers the smoothed functional average $\mu(t)$, we define a Functional Survival Discrimination Curve (FSDC) as follows:
\begin{equation}
\Tilde{X}_{\mathcal{A}_h}(t)=\mu(t)+ \sum_{h \in \Omega_h} \nu_{0h} \xi_h(t)\label{1}\qquad\mbox{with}\qquad t\in\mathcal{T}
\end{equation}
where  $\nu_{0h}$ represents the score's value selected within node $\mathcal{A}_h$ according to the splitting rule and $\Omega_h=\bigcup_{h\in H}\mathcal{A}_h$ denotes the set of nodes $\mathcal{A}_h$, where the eigenfunctions and FPCs' score values are selected based on the classification rule path in the FST. 

In FST, $\nu_{0h}$ can be seen as a threshold value selected by splitting methods at $h$-th node to differentiate survival curves (Keplan-Meier and so on) across groups. Hence,  at each newly reached node $\mathcal{A}_h$, the set of $\{\nu_{0h}\}_{h\in\Omega}$ represents the weight assigned to the FSDC at each step of separation. These values show how the FSDC changes over time. The critical curve acts as a reference threshold for potential state of survival changes. Beyond this threshold, an individual's state may undergo significant changes, marking a transition or critical event in the survival process. In our case, the FMST considers the mixed feature $\bm{W}_i=(x_{i1},\ldots,x_{iq},\hat{\nu}_{i\,q+1},\ldots,\hat{\nu}_{i\,q+p})$ for the $i$-th unit, the FSDC defines two half-spaces in $\mathcal{L}_2(\mathcal{T})$ at the $\mathcal{A}_h$ node given by:
\begin{align}
&R_1^h=\{\hat{X}_i(t)\,:\,w_{ih}\leq c \, ,\,i\in \mathcal{I}_{h} \} \notag\\& R_2^h=\{\hat{X}_i(t)\,:\,w_{ih}> c\, , \,i\in \mathcal{I}_{h}\}
\end{align}
with $w_{ih}\in\{\hat{\nu}_{ih},x_{ih}\}$ and $c\in\{\nu_{0h},x_{0h}\}$, where $x_{0h}$  value determined by a generic classical covariate $\bm{X}\in\mathbb{R}^q$ while $\nu_{0h}$ is the value selected by a FPC score. The set of approximated functions $\{\hat{X}_i(t)\}_{i\in\mathcal{I}_h}$ is computed as in Equation \ref{eq:predicted_trajectory} with $t\in \mathcal{T}$. Here, $I_{h}$ represents the set of statistical units assigned to daughter node $\mathcal{A}_{h}$ based on the classification path followed from the parent node $\mathcal{A}_{h-1}$. 

In $\mathcal{L}_2(\mathcal{T})$, at each node reached, a new discrimination curve is defined, allowing to visualise the survival functions into two distinct regions, $R_1^h$ and $R_2^h$, based on the values of the score $\bm{W}_{i}\in\mathbb{R}^{q+p}$ for $i\in\mathcal{I}_{h}$ selected in $\mathcal{A}_h$. 

However, the set of critical values $\{w_{ih}\}_{h \in \Omega_h}$, which are of varying nature, affects the shape and development of FSDC. Therefore, it is necessary to introduce a LFSDC that accounts for the weights introduced by both continuous and non-functional variables selected within each node of FMST. Firstly, we define the local mean function within a generic $h$-th node as $\hat{\mu}_{\mathcal{A}_h}(t)$ given by:
\begin{equation}
\hat{\mu}_{\mathcal{A}_h}(t)=\frac{1}{N_h}\sum_{i\in\mathcal{I}_h}\hat{X}_i(t)
\end{equation}  
where $N_h$ represents the number of statistical units assigned to the node $\mathcal{A}_h$, that is $N_h = |\mathcal{I}_h|$, while the term $\hat{X}_i(t)$ refers to the estimated trajectory for the $i$-th unit $i\in \mathcal{I}_h$ selected according to the splitting methods. The $\hat{\mu}_{\mathcal{A}_h}(t)$ considers the statistical units involved at a specific node $h$ according to the classification rule path and represents a conditional mean concerning the previous levels of the tree. When the $\hat{\mu}_{\mathcal{A}_h}(t)$ is used, we can observe how the local behaviour of the group of survival functions evolves at each depth in each node of the FMST. Thus, we introduce the LFSDC at the node $\mathcal{A}_h^*$ of the tree as follows:
\begin{equation}
\tilde{X}^{\text{Loc}}_{\mathcal{A}_h^*}(t) =  
\hat{\mu}_{\mathcal{A}_h}(t) + \sum\limits_{h\in \Omega_{h}^*} \nu_{0h} \xi_h(t)\quad\text{with}\quad t\in \mathcal{T} \label{eq:LFSDC}
\end{equation}
The LFSDC is conditioned by the selected variable in $\mathcal{A}_h$ and is adjusted based on contributions from the survival curves associated with the node. We denote by $\mathcal{A}_h^{**}$ the $h$-th node generated by selecting the classical $q$-th feature through splitting methods, while with $\mathcal{A}_h^*$ the $h$-th node obtained by selecting the $p$-th FPC's score as feature. The threshold value $c$ is assigned at each node differently depending on the situation. Thus, we define as $\Omega_{h}^* = \mathcal{A}_1^* \cup \mathcal{A}_2^* \cup \ldots \cup \mathcal{A}_{h}^*$, the set of threshold functional values $\{\nu_{0h}\}_{h\in \Omega^*_h}$ assigned at the $h$-th pass where the FPC's score is selected. These values quantify the contribution of the current node, while the local mean reflects changes in the survival state according to the nature of the variable, classical or functional, selected at the previous split.

The LFSDC shows how the impact of both variable selection at each node and the hierarchical structure influences the overall survival state, with each node's contribution weighted appropriately. The FMST is built by associating each node $\mathcal{A^*}$ with the LFSDC curve defined in Equation \ref{eq:LFSDC} along with the corresponding p-value obtained from the formula of the log-rank test given by Equation \ref{logrank}. Moreover, let $\tilde{X}^{Loc}_{\mathcal{A}_{h}^*}$ and $\tilde{X}^{Loc}_{\mathcal{A}_{h-1}^*}$ be the two LFSDC curves belonging to the space $\mathcal{L}_2(\mathcal{T})$, we define a metric to track changes in the survival state between successive localised nodes as follows: 
\begin{equation}
d_2(\tilde{X}^{Loc}_{\mathcal{A}_{h}^*},\tilde{X}^{Loc}_{\mathcal{A}_{h-1}^*}) =\Bigg(\int_{\mathcal{T}}|\tilde{X}^{Loc}_{\mathcal{A}_{h}^*}(t)-\tilde{X}^{Loc}_{\mathcal{A}_{h-1}^*}(t)|^{2}dt\Bigg)^{\frac{1}{2}}\,.
\label{eq:dist_sep}
\end{equation}
This metric quantifies the dissimilarity between the two LFSDC curves over the time domain $\mathcal{T}$, effectively capturing how the survival structure evolves from one node to the next.  
If $d_2(\tilde{X}^{Loc}_{\mathcal{A}_{h}^*},\tilde{X}^{Loc}_{\mathcal{A}_{h-1}^*})$ is large, it indicates that the two curves differ substantially, suggesting a significant shift in survival behavior between the parent and current node. Conversely, a small value implies a minor change, meaning the node retains a similar survival structure to its predecessor.

In the FMST, the graphical tools are essential to interpret and analyze the model’s results. These visual aids help clarify the survival dynamics at different nodes and provide an intuitive understanding of how functional covariates contribute to survival patterns. Specifically, at the terminal nodes $\{\mathcal{A}_s\}_{s\in\mathcal{S}}$, we characterize the graphical interpretability of $\mathrm{T}$ through the following graphical survival set:
\begin{equation}
    G_{\mathcal{A}_s}=\{\hat{X}_i(t)_{| i\in \mathcal{I}_{s}},\hat{H}_{\mathcal{A}_s}(t),\hat{S}_{\mathcal{A}_s}(t)\}
\end{equation}
where $\hat{X}_i(t)_{|i\in \mathcal{I}_s}$ represents the estimated trajectories of the selected covariate over time for all individuals $i\in\mathcal{I}_s$, while $\hat{H}_{\mathcal{A}_s}(t)$ and $\hat{S}_{\mathcal{A}_s}(t)$ refer to the cumulative hazard function and the survival function estimated by Equation \ref{estimators}.

The key steps of the method can be summarised as follows:
\begin{itemize}
    \item \emph{Step 1 - Data Preparation and Preprocessing}: Prepare the mixed survival dataset $D_N = \{(Y_i, \bm{X}_i, \bm{\hat{V}}_i, \bm{T}_i, \delta_i) : i = 1, \dots, N\}$ to construct $\mathrm{T}$;
    \item \emph{Step 2 - Recursive Tree Building}: Build the FMST recursively by splitting the dataset $D_N$ at each node to maximise the separation of survival distributions across resulting groups. At each node $\mathcal{A}_h$, apply the standard splitting methods (such as Log-rank, Log-rank score, etc.) based on the chosen feature or FPC score;
    \item \emph{Step 3 - Define a Local Functional Survival Discrimination Curve (LFSDC) at Each Node}: Define the LFSDC at each node $\mathcal{A}_h^*$ reflecting the local dynamics at that level of the tree;
    \item \emph{Step 4 - Computation of Separability Metric}: Compute a metric to track separability between successive localised nodes. This metric helps evaluate the effectiveness of splits at each step in separating the survival distribution;
    \item \emph{Step 5 - Terminal Node Interpretation}: At each terminal node $\mathcal{A}_s$, visually assess the functional curves and the survival risk for the group. This interpretation helps identify the key features that explain the differences in survival across groups and provides insights into clinical decision-making.
\end{itemize}
These steps collectively form a comprehensive framework for analysing and interpreting FSMT with both classical and functional covariates, enhancing both the predictive accuracy and the interpretability of the model in the functional space $\mathcal{L}^2(\mathcal{T})$.

\subsection{Explainability for Functional Random Survival Forest (FRSF)}

In machine learning, a common issue is understanding how the prediction model is constructed. When the internal mechanisms are not transparent, we refer to these as black-box models.
The Random Survival Forest (RSF) framework enables the evaluation of feature relevance through measures such as Variable Importance (VIMP), which quantify how each variable contributes to the model’s predictive performance. However, when dealing with heterogeneous features, such as continuous covariates, functional principal component (FPC) scores, or time-dependent variables, standard importance measures may be affected by scale differences, variable types, or temporal effects. These issues can distort the interpretation of a feature’s true impact on survival outcomes.

We propose a dual-level explainability framework for the Functional Random Survival Forest (FRSF) to address these limitations, incorporating global and local perspectives. Global explainability captures the overall influence of each variable across the entire ensemble of FSTs, helping to identify consistently important predictors. On the other hand, local explainability reveals how specific features affect individual predictions over time, offering personalised insights and highlighting dynamic interactions that may not emerge from global measures alone.
This combined approach enhances transparency and provides a more nuanced understanding of how functional and classical covariates influence survival trajectories, particularly in models where temporal complexity and heterogeneity play a central role.

\subsubsection{Local Method: Survival Shapley Additive Explanations (\textit{Surv}SHAP)}

Let $\mathscr{L}_N = \{(\bm{W}_1, Y_1), \dots, (\bm{W}_N, Y_N)\}$ be the learning data set where $\bm{W}_i\in\{\bm{X}_i,\bm{\hat{V}}_i\}$ ($\bm{W}_i\in \mathbb{R}^{q+p}$), for $i=1,\ldots,N$, represents the set of features that involves both the non-functional feature set and the additional features derived by PACE decomposition. This combined feature set is used to construct the FRSF model. 

The \textit{Surv}SHAP$(t)$ algorithm is a method designed for time-dependent explanations for a survival model \citep{KRZYZINSKI2023110234}. In general, in survival analysis assuming that $t_1,\ldots,t_M$ are distinct times to event of interest for each subject $i\in\mathcal{I}$, the FRSF returns the estimated $\mathcal{\hat{S}}(t|\bm{w})$ for a given response $\bm{W}=\bm{w}$. The algorithm calculates the \textit{Surv}SHAP$(t)$ for an individual observation $\bm{w}_*$, given by functions  $[\phi_{t_1}(\bm{w}_{*},j),\phi_{t_2}(\bm{w}_{*},j),\ldots,\phi_{t_M}(\bm{w}_{*},j)]$ generated for each component $j\in\{1,\ldots,q+p\}$. These functions quantify the time-dependent effects of the features on the model's predictions, illustrating how the contributions of each variable evolve and influence the survival outcomes over time. In \textit{Surv}SHAP$(t)$ algorithm the contribution of the $j$-th component in the time $t_m$ is given by:
\begin{equation}
\phi_{t_m}(\bm{w}_{*},j)=\frac{1}{|\Pi|}\sum_{\pi\in\Pi}e^{before\,(\pi,j)\cup j}_{t_m,\bm{w}_*}-e^{before\,(\pi,j)}_{t_m,\bm{w}_*}
\label{survshap}
\end{equation}

\noindent where $|\Pi| = (q + p)!$ represents the total number of permutations of the $q+p$ components. The term $e^{before\,(\pi,j)\cup j}_{t_m, \bm{w}_*}$ denotes the expected survival prediction at time $t_m$, since the $j$-th component has been included in the subset of the components that occur before $j$ in the permutation $\pi$. Instead, $e^{before\,(\pi,j)}_{t_m, \bm{w}_*}$ represents the expected prediction based on the subset of characteristics occurring before $j$, without including $j$ itself. Thus, the term $e^{\,C}_{t,\bm{w}_*} = E[\hat{\mathcal{S}}(t| \bm{w}) | \bm{w}^{C} = \bm{w}_{*}^{C}]$ represents the expected survival function $\hat{\mathcal{S}}(t,\bm{w})$ conditioned on the values of the subset $C$ of the components of the feature. 

To make it easier to compare different models and time points, this value can be normalised to a scale \([-1,1]\) as follows:
\begin{equation}
  \phi_{t_m}^*(\bm{w}_{*},j)=\frac{\phi_{t_m}(\bm{w}_{*},j)}{\sum_{d=1}^{q+p}|\phi_{t_m}(\bm{w}_{*},d)|}\label{survshap_norm}
\end{equation}

Another way to calculate the \textit{Surv}SHAP$(t)$ functions is the Shapley kernel model (ref) and performing the weighted linear regression with functional responses and scalar variables. Let the binary response  $\bm{z}_d \in \{0,1\}^{q+p}$ for $d\in\{1,\ldots,D\}$ be a binary vector that indicates whether a specific component is included in the model, with $1$ representing inclusion and $0$ exclusion. Moreover, given a function $h_{\bm{w}}:\{0,1\}^{q+p}\rightarrow \mathbb{R}^{q+p}$ which converts binary vectors into original input for $z_d=1$ in $d$-th element, the  \textit{Surv}SHAP$(t)$  is calculated as follows:
\begin{equation}
\bm{\phi}=(\bm{Z^TWZ})^{-1}\bm{Z^TWY}\,.
\end{equation}
The $\bm{Z}$ is a matrix of binary responses, while $\bm{W}$ is the diagonal matrix of the weights associated with each binary vector $\bm{z}$ given by:
\begin{equation}
    w(\bm{z})=\frac{(q+p)-1}{\binom{q+p}{k}k[(q+p)-k]}
\end{equation}
with $k$ the number of ones in the vector $\bm{z}$, and $\bm{Y}$ is a matrix of the survival function values obtained from the mapping $h_{\bm{w}}$.

Although \textit{Surv}SHAP$(t)$ has been proposed for time-dependent local explanations in survival models, it has never been extended to handle settings with mixed (functional and scalar) predictors, nor to treat the resulting contributions as functional objects. Incorporating an FDA perspective allows us to manage the complex nature of both inputs and explanations in a coherent and smooth temporal framework.
We propose to approximate the time-dependent contribution of the $j$-th component for an individual $i$ the \textit{Surv}SHAP$(t)$ as a smooth function over time $\mathcal{T}$. Thus defined as $\bm{\Phi}_j^*=(\phi_{1j}^*,\phi_{2j}^*,\ldots,\phi_{Mj}^*)$ the $j$-th time survival shapley vector, where $\phi_{m\,,\,j}^*=\phi_{t_m}^*(\bm{w}_{*},j)$ calculated from Equation \ref{survshap_norm} for $m=1,\ldots,M$, the full \textit{Surv}$\text{SHAP}_{ij}(t)$ function can be obtained as a problem of minimisation given by:
\begin{equation}
 \arg\min_{f\in\mathcal{F}} \sum_{m=1}^M \left(\phi_{m,j}^* - f(t_m)\right)^2 + \lambda \int_{\mathcal{T}} \left(\frac{d^2}{dt^2} f(t)\right)^2 dt, \label{smooth}
\end{equation}
where $\lambda$ is a regularisation parameter that controls the smoothness of the solution, the second term is the regularisation term, which penalises large variations in the second derivative of the fitted function $f(t)$ chosen in the functional space $\mathcal{F}$ over the period $\mathcal{T}$ for the $j=1,\ldots,q+p$ and $i=1,\ldots,N$.
Additionally, we obtain the local time information for $j$-th contribution to the $i$-th unit in the FRSF model over time $\mathcal{T}$ as follows:
\begin{equation}
\Delta_{t_{\alpha},\,t_{\beta}}^{ij}=\int_{t_{\alpha}}^{t_{\beta}}\textit{Surv}\text{SHAP}_{ij}(\tau)\,d\tau\label{delta:survshap}
\end{equation}
where $t_{\alpha}$ and $t_{\beta}$ represent the boundary times within which no events occur within time interval $\mathcal{T}$. This integral calculates the accumulated contribution of the $j$-th feature in the model over the specified time range between $[t_{\alpha},t_{\beta}]$, providing a local measure of the effect for the $j$-th component. We then define the Time SHAP Difference (TSD) as the difference between the local accumulated contribution \(\Delta_{t_{\alpha}, t_{\beta}}^{ij}\) and the aggregated local contribution \(\Gamma_{t_{\alpha}, t_{\beta}}^{ij}\), where \(\Gamma_{t_{\alpha}, t_{\beta}}^{ij}=\int_{t_\alpha}^{t_\beta} \phi_{t}^*(\bm{w}_{*},j)\,dt\) represents the standard \textit{Surv}SHAP$(t)$ based aggregate contribution in the classical setting. This difference allows us to quantify the temporal variation in the feature's importance compared to its conventional static counterpart:

\begin{equation}
\text{TSD}_{t_{\alpha}, t_{\beta}}^{ij} = \Delta_{t_{\alpha}, t_{\beta}}^{ij} - \Gamma_{t_{\alpha}, t_{\beta}}^{ij} \label{TSD}
\end{equation}

This measure quantifies how much the time-dependent nature of the model influences the $j$-th feature's impact within the specified time range \([t_{\alpha}, t_{\beta}]\). If \(\text{TNSD}_{t_{\alpha}, t_{\beta}}^{ij}>0\), then the accumulated time-dependent contribution of the feature is greater than its static contribution. This means that the feature effects over time \([t_{\alpha}, t_{\beta}]\) have a larger time-dependent influence than the one obtained classical value of \textit{Surv}SHAP$(t)$ values. If \(\text{TNSD}_{t_{\alpha}, t_{\beta}}^{ij}<0\), this suggests that the feature has less time-dependent influence than the static \textit{Surv}SHAP$(t)$ values. These \(\text{TNSD}_{t_{\alpha}, t_{\beta}}^{ij} \) can help identify features with varying importance depending on the time. However, we can define the normalised form of Equation \ref{TNSD} defining the Time Normalised SHAP Difference (TNSD) as follows:
\begin{equation}
\text{TNSD}_{t_{\alpha}, t_{\beta}}^{ij} = \frac{\Delta_{t_{\alpha}, t_{\beta}}^{ij} - \Gamma_{t_{\alpha}, t_{\beta}}^{ij}}{t_\beta-t_\alpha} \label{TNSD} 
\end{equation}

This measure captures the extent to which the model's time-dependent structure affects a given feature's relevance over a specific time interval. A positive value indicates that the feature has a greater influence when its effect is considered dynamically over time, compared to a static perspective. In contrast, a negative value suggests that the feature plays a less significant role when considering temporal variation.
By expressing this difference in a normalised form, relative to the length of the interval, it becomes possible to interpret it as a rate of change in the feature's importance. This normalization is beneficial as it allows for fair comparisons across time intervals of different lengths, helping to identify features whose influence varies more or less over time.

\subsubsection{Global Method: Permutation Feature Importance (PFI)}
In survival analysis, a commonly used global method to assess feature importance is the Permutation Feature Importance (PFI) \citep{Mi2021}. Within the FRSF framework, PFI may be employed to evaluate the impact of the FPC's scores on the model’s predictions, and to compare their relevance with that of the traditional covariates used in classical RSF models.

Let $\hat{S}(\cdot)$ be the trained survival model generated by FRSF, $\bm{W}\in\mathbb{R}^{N\times (q+p)}$ the mixed feature matrix and $\delta\in\{0,1\}^N$ the censoring indicator, which captures the status of each observation in the dataset. The PFI in FRSF starts from the survival time vector $\bm{t}=(t_1,\ldots,t_M)$ and a fixed loss function $L(\bm{t},\hat{S}(\cdot))$ (such as the Brier score and so on). The method can be systematically computed to evaluate the importance of the significance variable as a change in the loss function after permutations of the variable values.

Initially, the algorithm calculates the baseline performance of the model using the original set of characteristics. This is represented mathematically as:
\begin{equation}
FI_0 = L(\bm{t}, \hat{S}(\bm{t},\bm{W})),  
\end{equation}
where $FI_0$ is the loss function associated with the unchanged feature matrix. 

To assess the impact of a particular feature on the model's predictions, we permute the $j$-th values of the matrix-vector given by $\bm{W}=(\bm{W}^{(1)},\bm{W}^{(2)},\ldots,\bm{W}^{(j)},\ldots,\bm{W}^{(q+p)})$ across all samples to generate a modified version denoted as $\bm{W}^{(j)}$. The performance of the model, with the permuted feature matrix $\bm{W}^{(j)}$, can then be computed:
\begin{equation}
FI_j = L(\bm{t}, \hat{S}(\bm{t},\bm{W}^{(j)}))  
\end{equation}
where $FI_j$ denotes the loss function incurred when the $j$-th feature has been permuted. 

The PFI for the $j$-th feature is calculated by comparing the baseline model performance to the performance after permuting the feature, as expressed by:

\begin{equation}
\text{FI}_j^{\,diff} = FI_0 - FI_j
\end{equation}

\noindent denoting the performance after permuting the $j$-th feature. This process is repeated for each $j = 1, \ldots, q + p$. A higher PFI value indicates a greater contribution of that feature to the model's predictive capacity.

This approach offers a robust method for evaluating feature significance, showing the role of the FPCs in survival predictions within the FRSF framework, thereby deepening our understanding of the model's decision-making process. Moreover, letting $\pi$ denote the number of permutations performed for the $j$-th component, we define the averaged  time importance score as:
\begin{equation}
\overline{\text{FI}}_j(t) = \frac{1}{\pi} \sum_{r=1}^{\pi} \Big(FI_0(t) - FI_{j}^{(r)}(t)\Big) \label{global_values}
\end{equation}
where \(FI_{j}^{(r)}(t)\) is the model performance at the \(r\)-th permutation of the $j$-th feature in time $t$, while $FI_0(t)$ is the baseline performance evaluated at same time $t$ . This average provides a more stable estimate of feature importance by mitigating the variability introduced by random permutations.

However, by considering the vector $\bm{\varrho}_j = (\varrho_{1j}, \varrho_{2j}, \ldots, \varrho_{Mj})$, where each element is defined as $\varrho_{mj} = \overline{\text{FI}}_j(t_m)$, we can represent the global temporal evolution of the importance for the $j$-th feature over the time $\mathcal{T}$ using a smooth function, as expressed in Equation \ref{smooth}. In this context, we define the global time-dependent contribution of the \( j \)-th feature within the FRSF model between two time points \( t_\alpha \) and \( t_\beta \) over the time interval \( \mathcal{T} \) as \( \Delta_{t_{\alpha},\,t_{\beta}}^{j} \), computed as in Equation \ref{delta:survshap}. This is done by applying a new smooth function to the values \( \overline{\text{FI}}_j(t_m) \) at each event time \( m = 1, \ldots, M \).
We define the Mean Time Global Difference (MTGD) and the Mean Time Normalized Global Difference (MTNGD) as follows, extending the concept used in the local case (\ref{TSD},\ref{TNSD}) for time-dependent contributions:

\begin{align}
&\text{MTGD}_{t_{\alpha}, t_{\beta}}^{j} = \Delta_{t_{\alpha}, t_{\beta}}^{j} - \Upsilon_{t_{\alpha}, t_{\beta}}^{j}\notag\\
&\text{MTNGD}_{t_{\alpha}, t_{\beta}}^{j} = \frac{\Delta_{t_{\alpha}, t_{\beta}}^{j} - \Upsilon_{t_{\alpha}, t_{\beta}}^{j}}{t_\beta-t_\alpha} 
\end{align}

Here, \( \Upsilon_{t_{\alpha}, t_{\beta}}^{j} \) represents the aggregated global contribution calculated based on \( \overline{\text{FI}}_j(t) \) at the time points \( t_\alpha \) and \( t_\beta \). This definition extends the local analysis framework to assess the global contribution of the \( j \)-th feature over time \( \mathcal{T} \).

\section{Application}
\subsection{Simulation study}

The proposed methods' performance was evaluated using simulated datasets designed to replicate real-world survival scenarios. This section systematically examines the effects of various interpretability-enhancing techniques by comparing standard STs and flexible FRSF models.

A comprehensive simulation study was conducted to assess survival outcomes across datasets with sample sizes \( N \in \{200, 300\} \) and follow-up durations \( |\mathcal{T}| \in \{180,100\} \). Time-dependent covariates were explicitly generated to evaluate their impact on survival. Two groups were generated, with 40\% of the units representing event occurrences and the remaining 60\% representing censoring that occurred. For each group of individuals \(k\in\{1,2\}\), the underlying trajectories functions \( f_{ik}(t) \) and \( g_{ik}(t) \), describe the survival function over time for the two sample sizes for $i=1,\ldots,N_k$. The models were defined for the groups of individuals as follows:
\begin{align}
    &f_{i1}(t) = a_{1i} \sin(\pi t) + a_{2i} \cos(\pi t)\, \notag \\
    &f_{i2}(t) = f_{i1}(t)\,(1 - u_i) + f_{i1}(t)\,u_i &  \quad t \in [0, 180] \label{first_180} \\
    &g_{i1}(t) = \mu t \notag \\
    &g_{i2}(t) = g_{i1}(t)+ksin(r\pi(t+\theta)) &  \quad t \in [0, 100] \label{second_200}
\end{align}
where \( a_{1i}\) and \(a_{2i}\) in Equation \ref{first_180} follow the uniform distribution within interval $[1,20]$ for $k=1$, while the intervals $[15,30]$ (first member) and $[10,30]$ (second member) for $k=2$ and $u_i\sim \mathcal{B}(0.60)$. For the Equation \ref{second_200} was selected, $\mu=4$, $k=12$, $r=0.7$ and $\theta\sim U[0.10,0.45]$. Moreover for each group was added an error \(e_i(t) \sim \mathcal{N}(0, K(s,t))\) where the covariance function is given by:
\begin{equation}
K(s,t) = 12\, e^{\left(-0.5 \,|t - s|\,\right)}\,.
\end{equation}
To model the observed time-dependent covariates for each unit \( i \) at each time \( t \in \mathcal{T}_i \), where \( \mathcal{T}_i \) represents the irregular time for the $i$th unit generated using a global time sample, we introduce an observation model that adds noise to the true underlying trajectories (\ref{first_180},\ref{second_200}) as follows:
\begin{align}
&Y_{i}(t_{ij}) = h_{ik}(t_{ij}) + \eta_{i}(t_{ij})\quad\text{with}\quad t_{ij}\in\mathcal{T}_i\notag\\ 
&\quad \text{for}\quad j=1,\ldots,J_i\quad i=1,\ldots,N_k\,.
\end{align}
where $h_{ik}(t)\in\{f_{ik}(t),g_{ik}(t)\}$ for $k=1,2$  and $\eta_{i}(t) \sim \mathcal{N}(0, 18)$. 
Moreover, categorical and numerical variables, $\bm{X}_i\in\mathbb{R}^4$ for $i=1,\ldots, N_k$, were generated using a Bernoulli distribution, while numerical variables were generated from a normal distribution.

In the first scenario, we generated the trajectories estimated for each sample size $N_k$ using PACE decomposition, selecting 
\(p=14\) as principal components to capture the underlying functional structure.  In functional classification settings, components explaining a minimal portion of the variability may still be crucial for learning subtle temporal patterns; thus, relying solely on variance-based selection criteria may lead to suboptimal model performance.

\begin{figure*}[tbp]
    \centering
    \begin{subfigure}{0.49\textwidth}
        \centering
        \includegraphics[width=\linewidth]{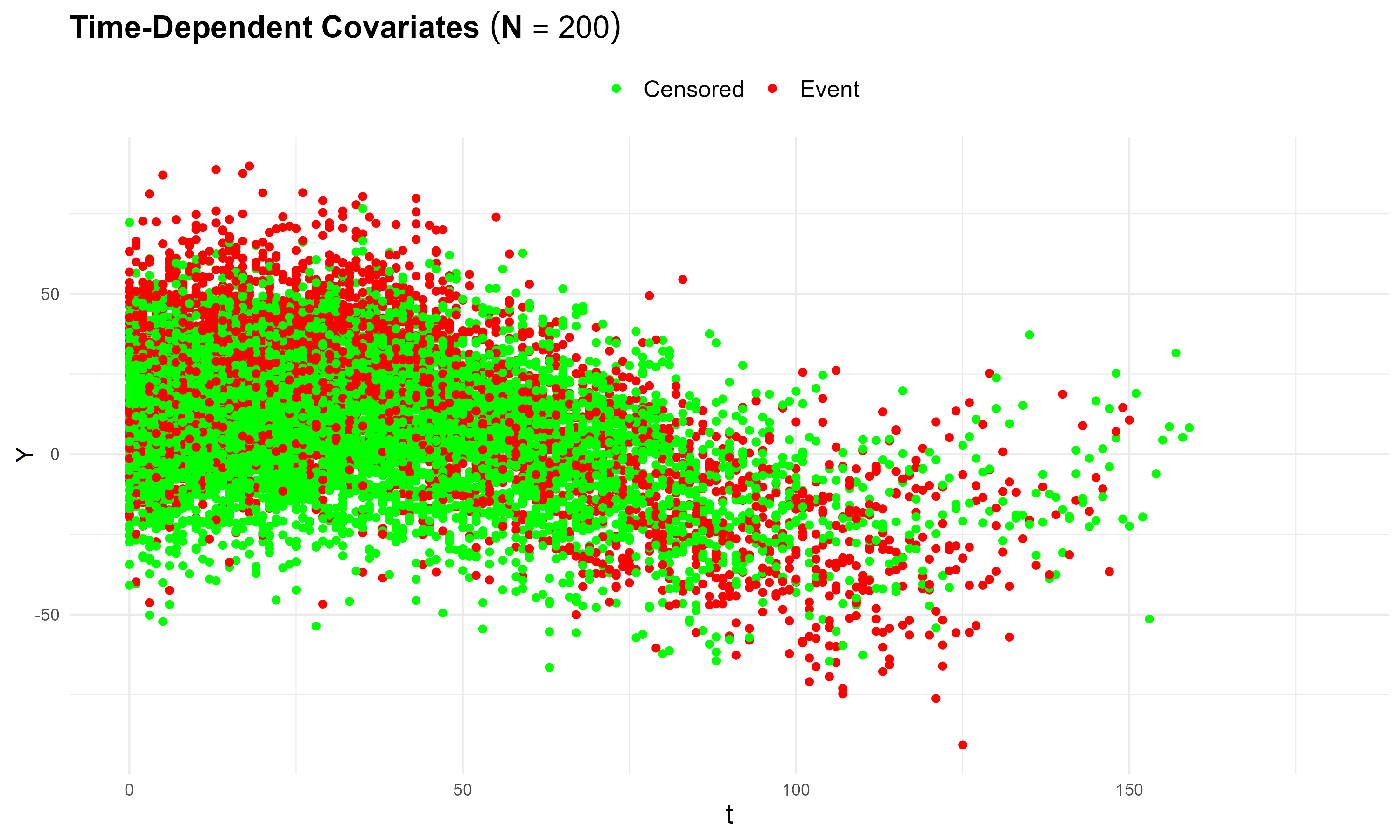} 
    \end{subfigure}
    \hfill
    \begin{subfigure}{0.49\textwidth}
        \centering
        \includegraphics[width=\linewidth]{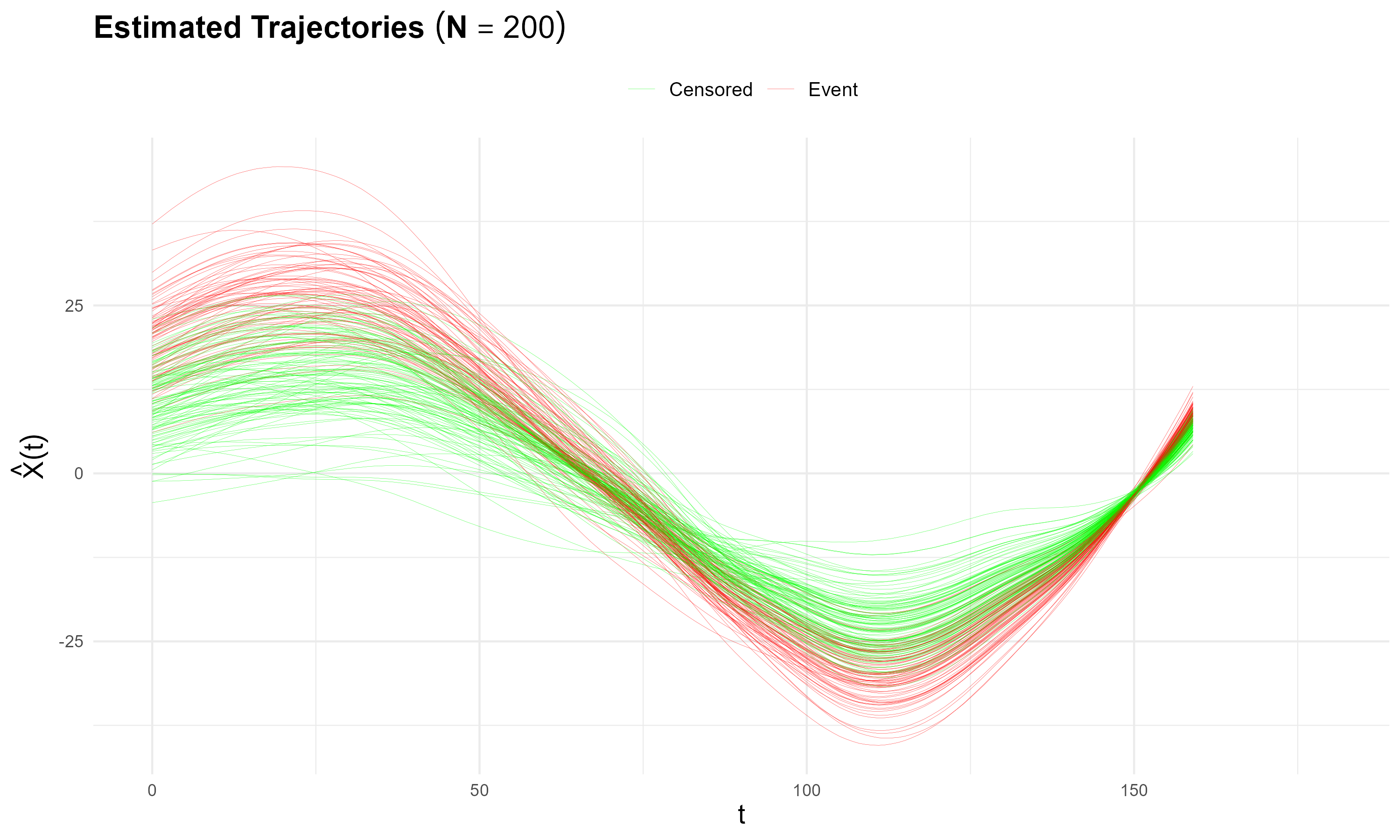} 
    \end{subfigure}
    \vfill

    \begin{subfigure}{0.49\textwidth}
        \centering
        \includegraphics[width=\linewidth]{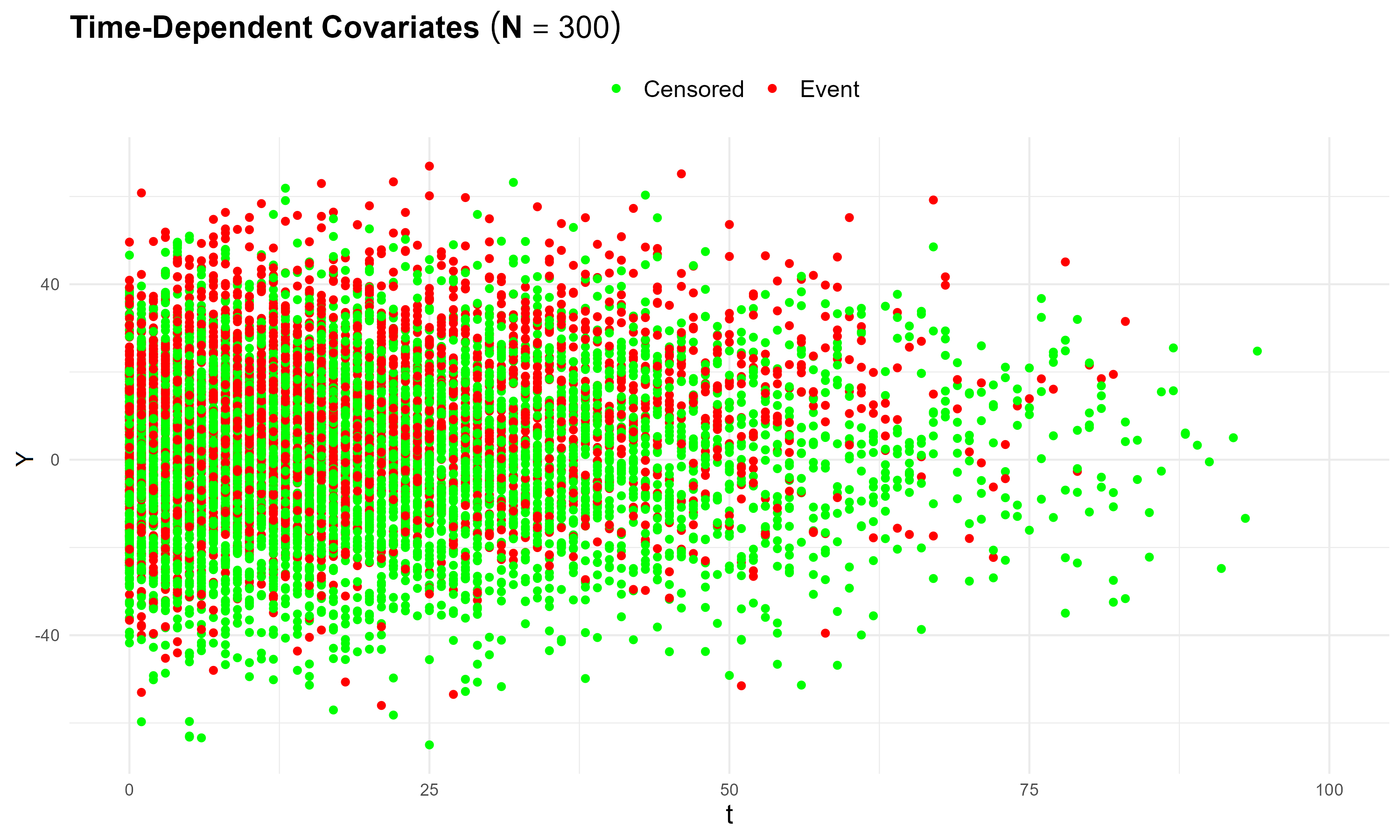}
    \end{subfigure}
    \hfill
    \begin{subfigure}{0.49\textwidth}
        \centering
        \includegraphics[width=\linewidth]{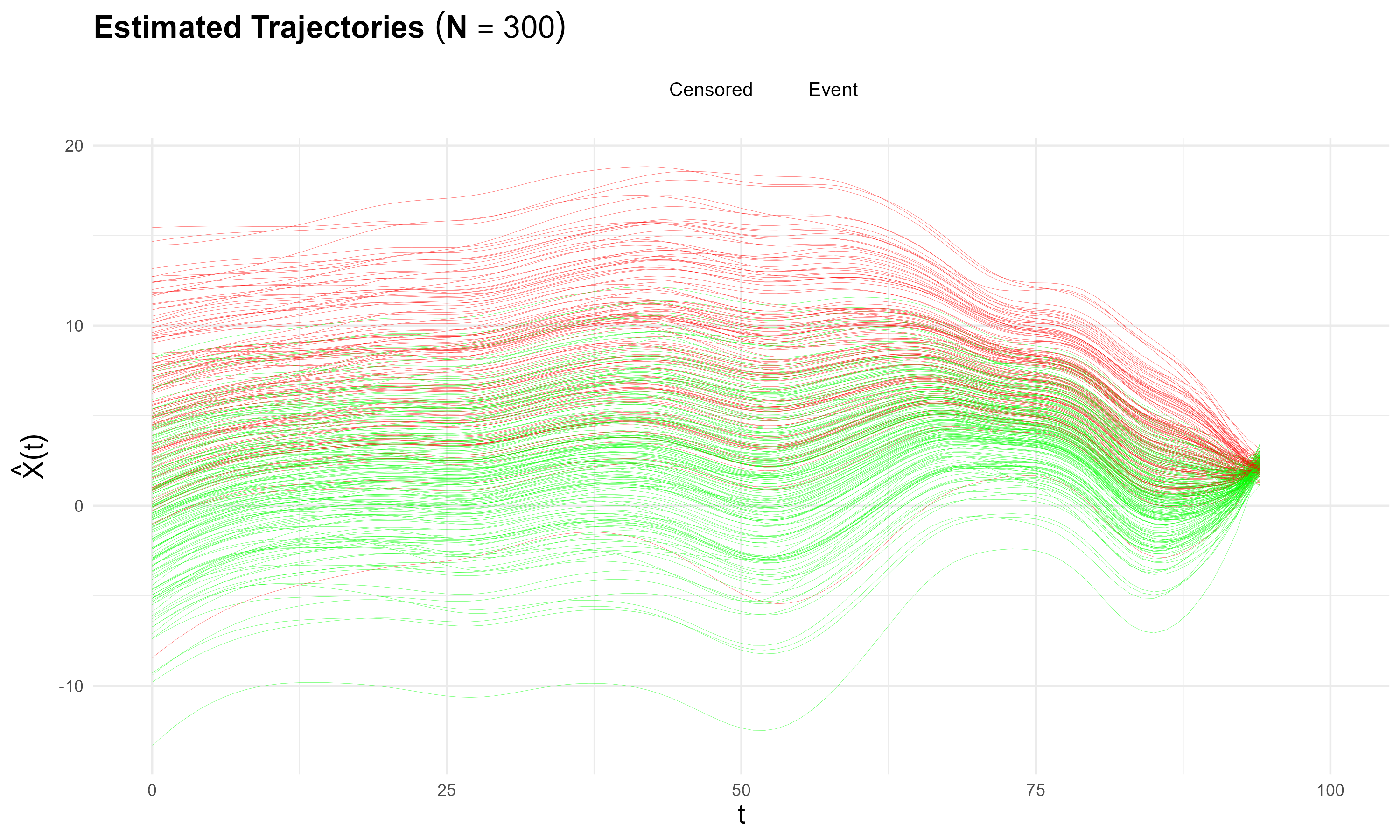}
    \end{subfigure}

    \caption{Visualization of time-dependent covariates and estimated trajectories for $N=200$ (Upper) and $N=300$ (Lower) stratified by censoring status and event occurrence.}
    \label{fig:data_simulated}
\end{figure*}

 Figure \ref{fig:data_simulated} illustrates the time-dependent data and estimated trajectories obtained through this method for the two cases. 

\begin{figure*}[tbp]
    \centering
    \begin{subfigure}{0.49\textwidth}
        \centering
        \includegraphics[width=\linewidth]{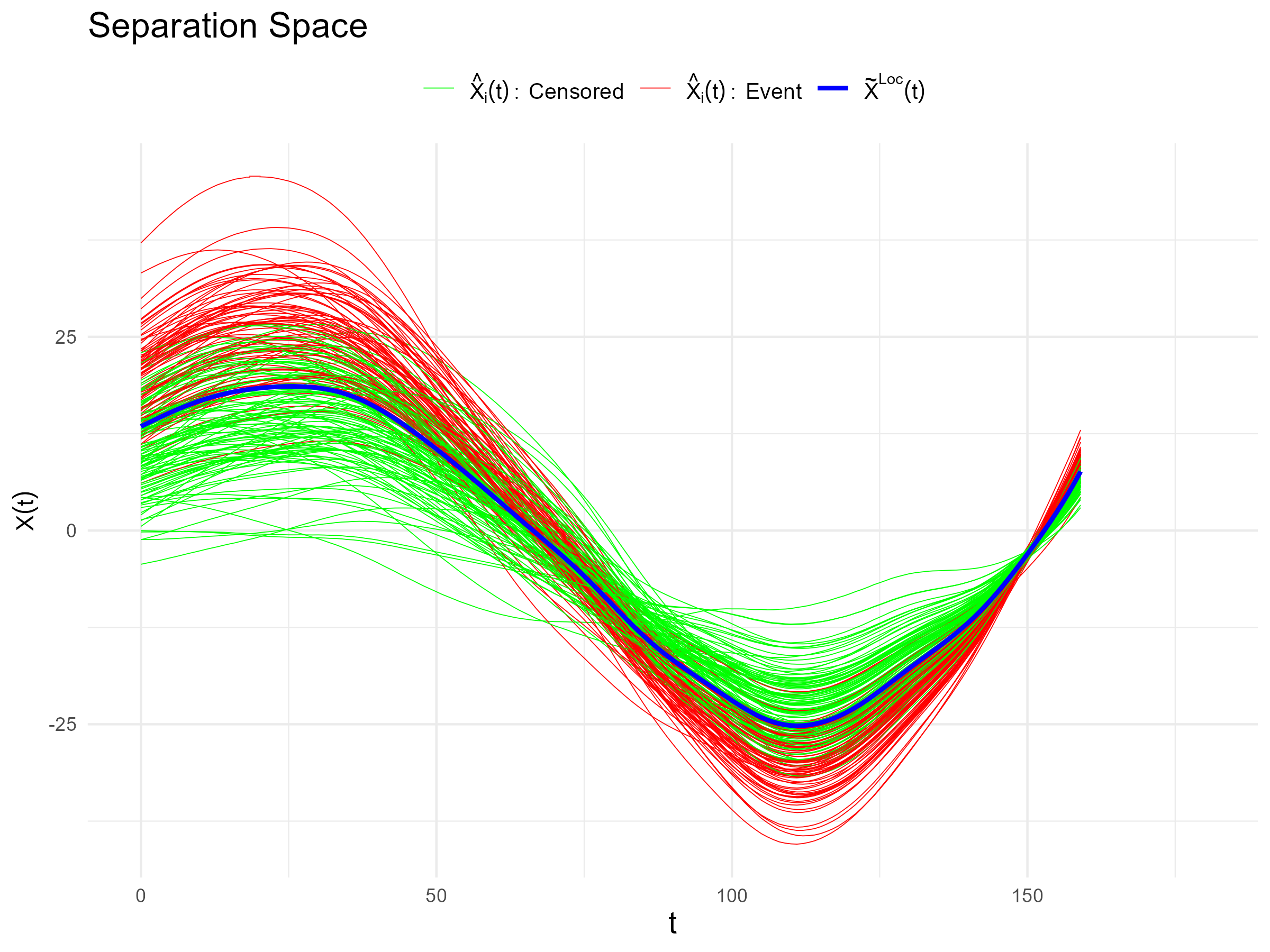} 
    \end{subfigure}
    \hfill
    \begin{subfigure}{0.49\textwidth}
        \centering
        \includegraphics[width=\linewidth]{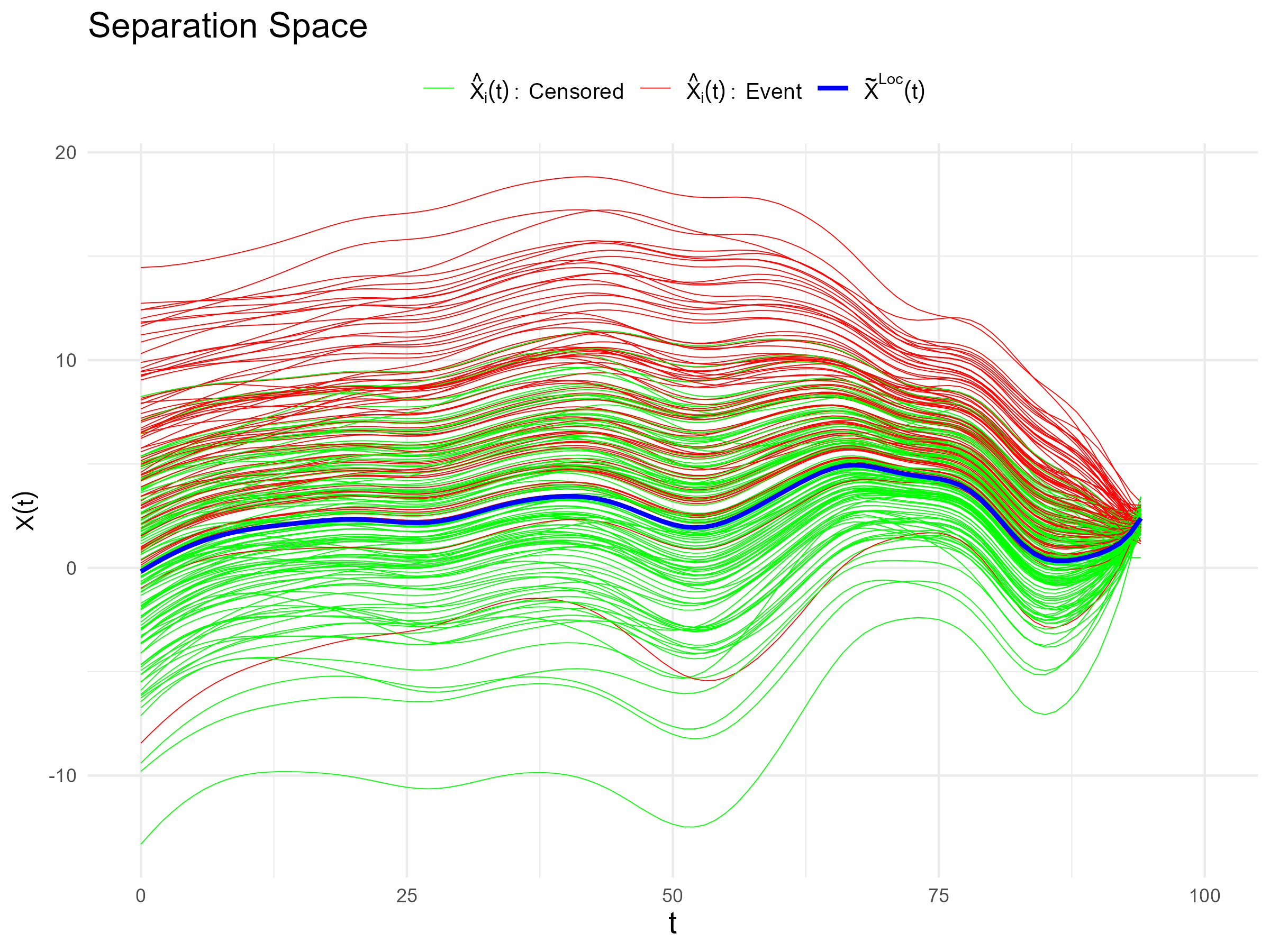} 
    \end{subfigure}
    
    \caption{Visualisation of the separation regions in the root node for two different sample sizes, \(N = 200\) (left) and \(N = 300\) (right).}   
    \label{fig:sep_simulated22}
\end{figure*}

The second scenario focused on determining FMST, aiming to investigate the graphical interpretability of the tree structure in a functional context. We explored how the functional covariates influenced the splitting criteria at each decision node, leading to the construction of distinct separation spaces. Figure \ref{fig:sep_simulated22} highlights the identified separation regions and the corresponding LFSDC (Blue function) in the root node.   

\begin{figure*}[tbp]
    \centering
    \begin{subfigure}{0.49\textwidth}
        \centering
        \includegraphics[width=\linewidth]{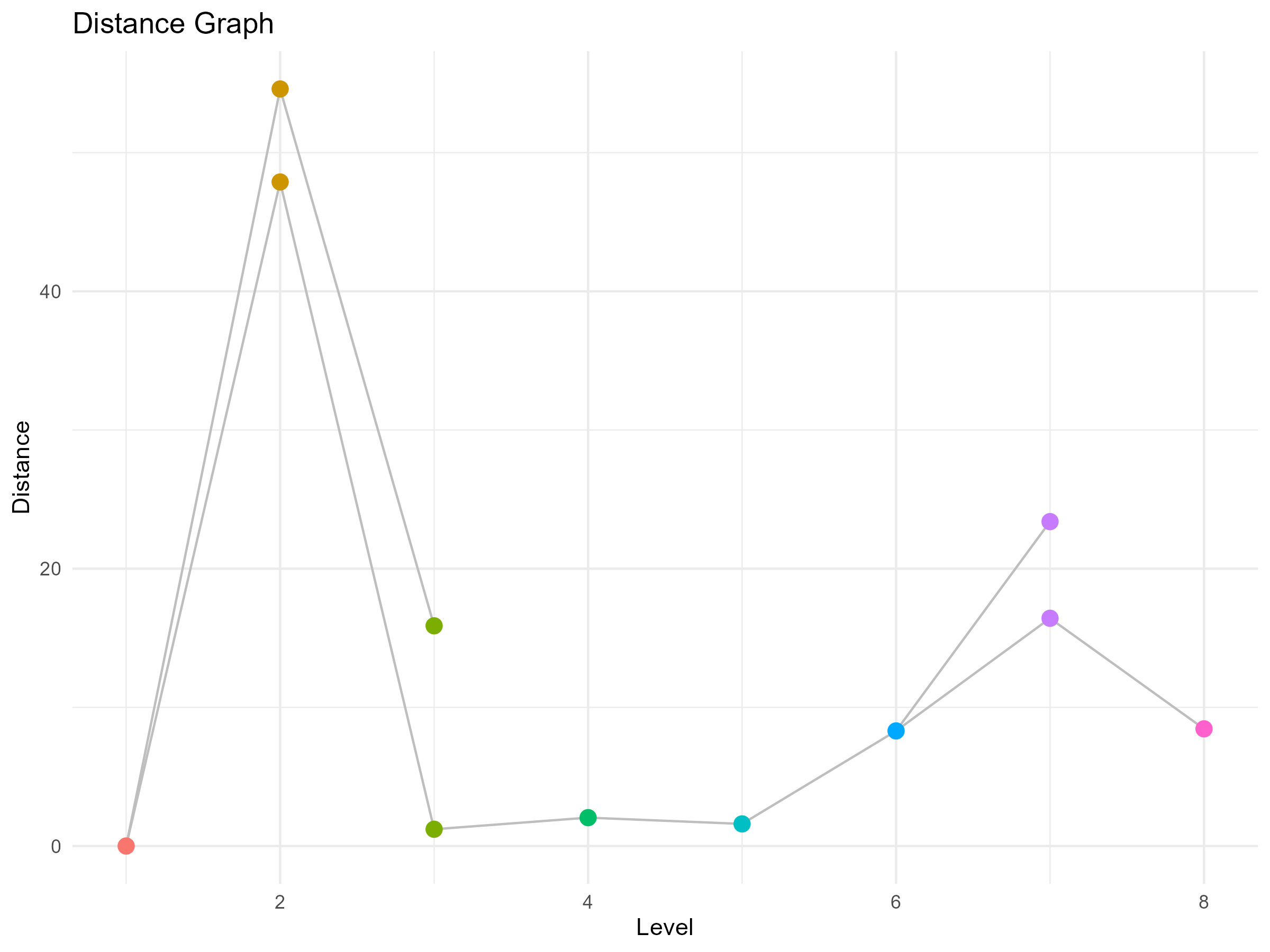} 
    \end{subfigure}
    \hfill
    \begin{subfigure}{0.49\textwidth}
        \centering
        \includegraphics[width=\linewidth]{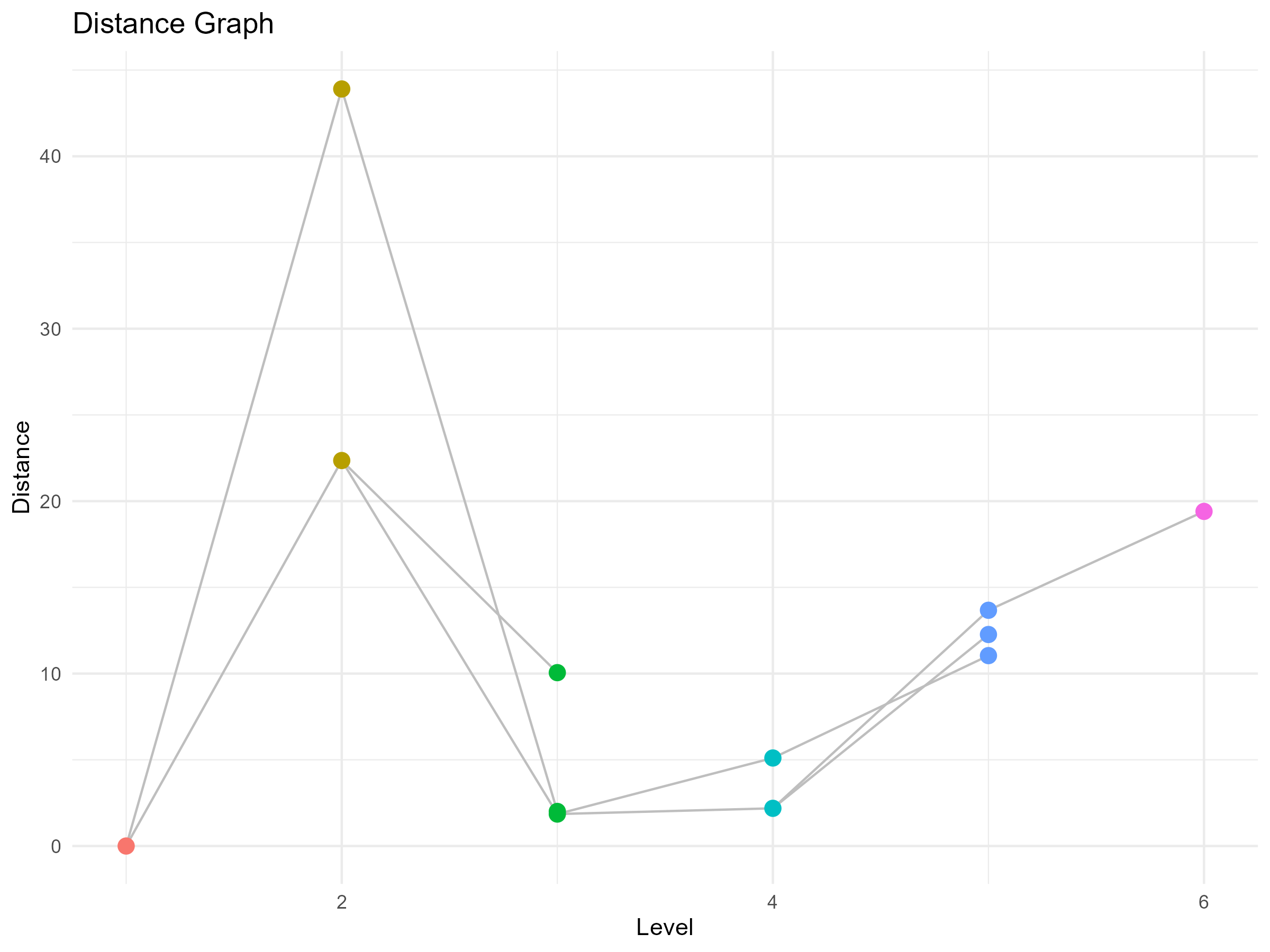} 
    \end{subfigure}
    
    \caption{Survival dynamics of distance at each level for $\mathcal{A}^*_h$ nodes of FMST for $N=200$ (left) and $N=300$ (right)}  \label{fig:distance}
\end{figure*}

Furthermore, we examined the overall tree structure by analyzing the functional survival dynamics describing each node's transition process. We monitored the separation process at each level by measuring the distance between the parent node and its daughter nodes using Equation \ref{eq:dist_sep}. This allowed us to assess the survival process of separation at each level as shown in Figure \ref{fig:distance}. 

\begin{figure*}[tbp]
    \centering
    \begin{subfigure}{0.49\textwidth}
        \centering
        \includegraphics[width=\linewidth]{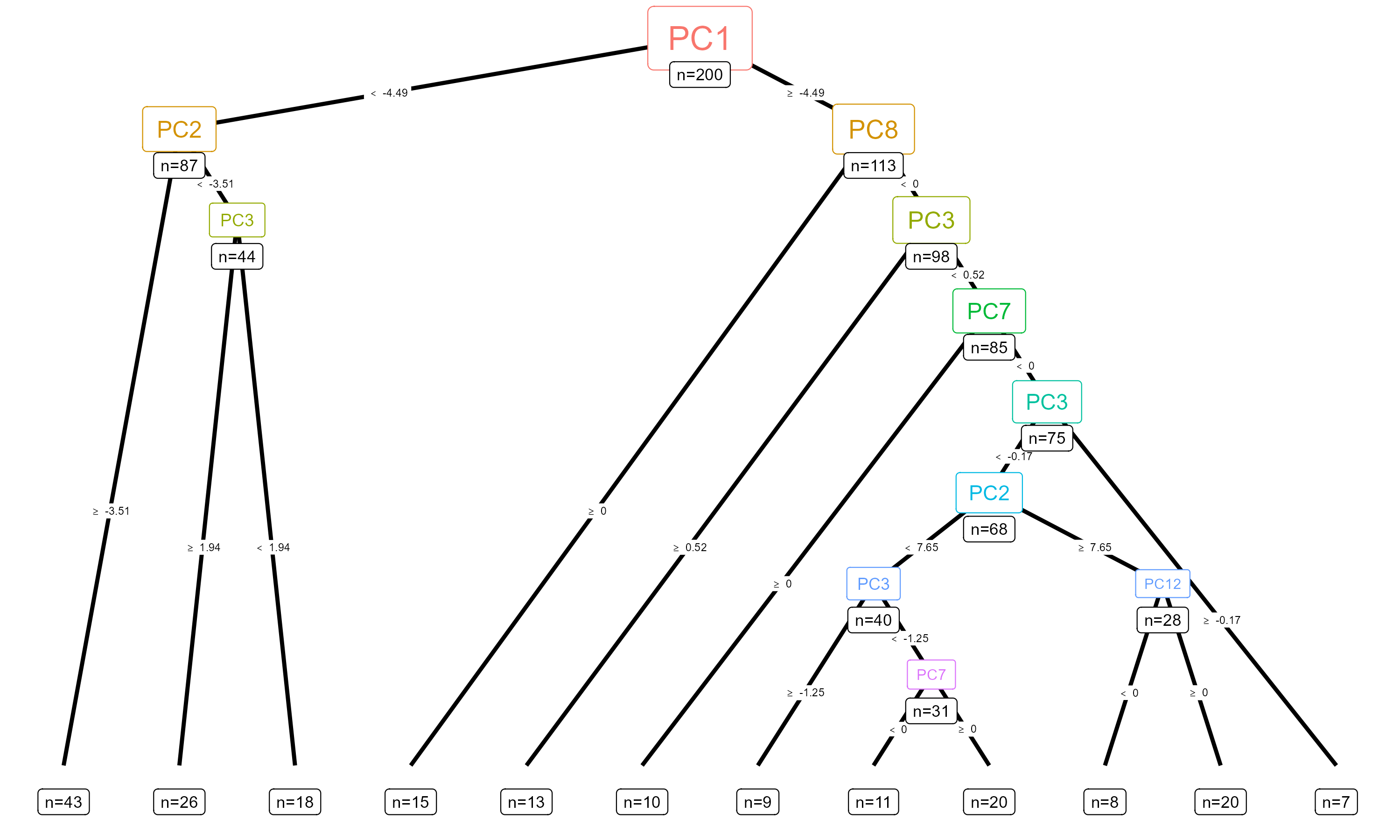} 
    \end{subfigure}
    \hfill
    \begin{subfigure}{0.49\textwidth}
        \centering
        \includegraphics[width=\linewidth]{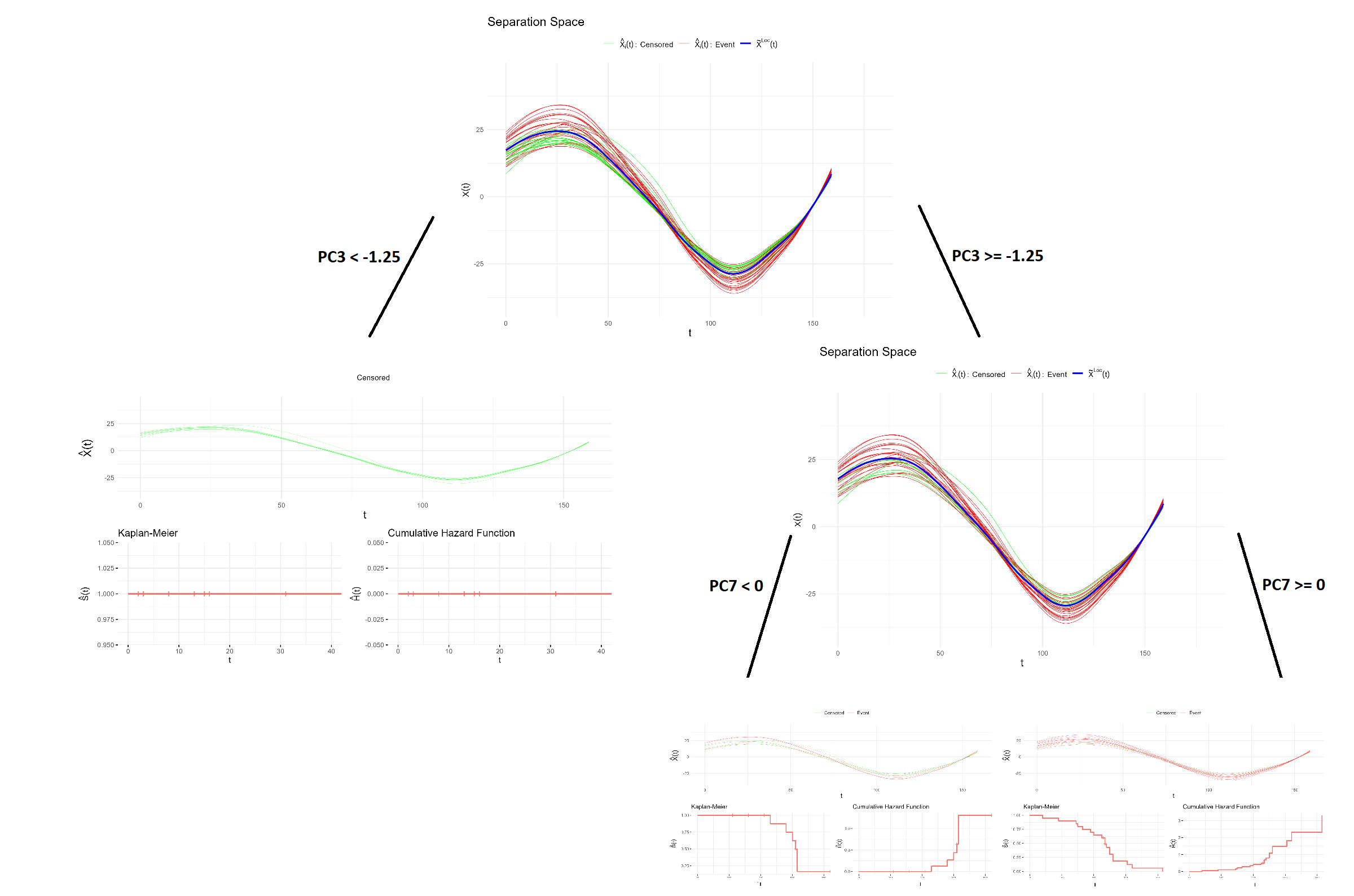} 
    \end{subfigure}
    \label{fig:sep_simulated}
    \vfill
    \begin{subfigure}{0.49\textwidth}
        \centering
        \includegraphics[width=\linewidth]{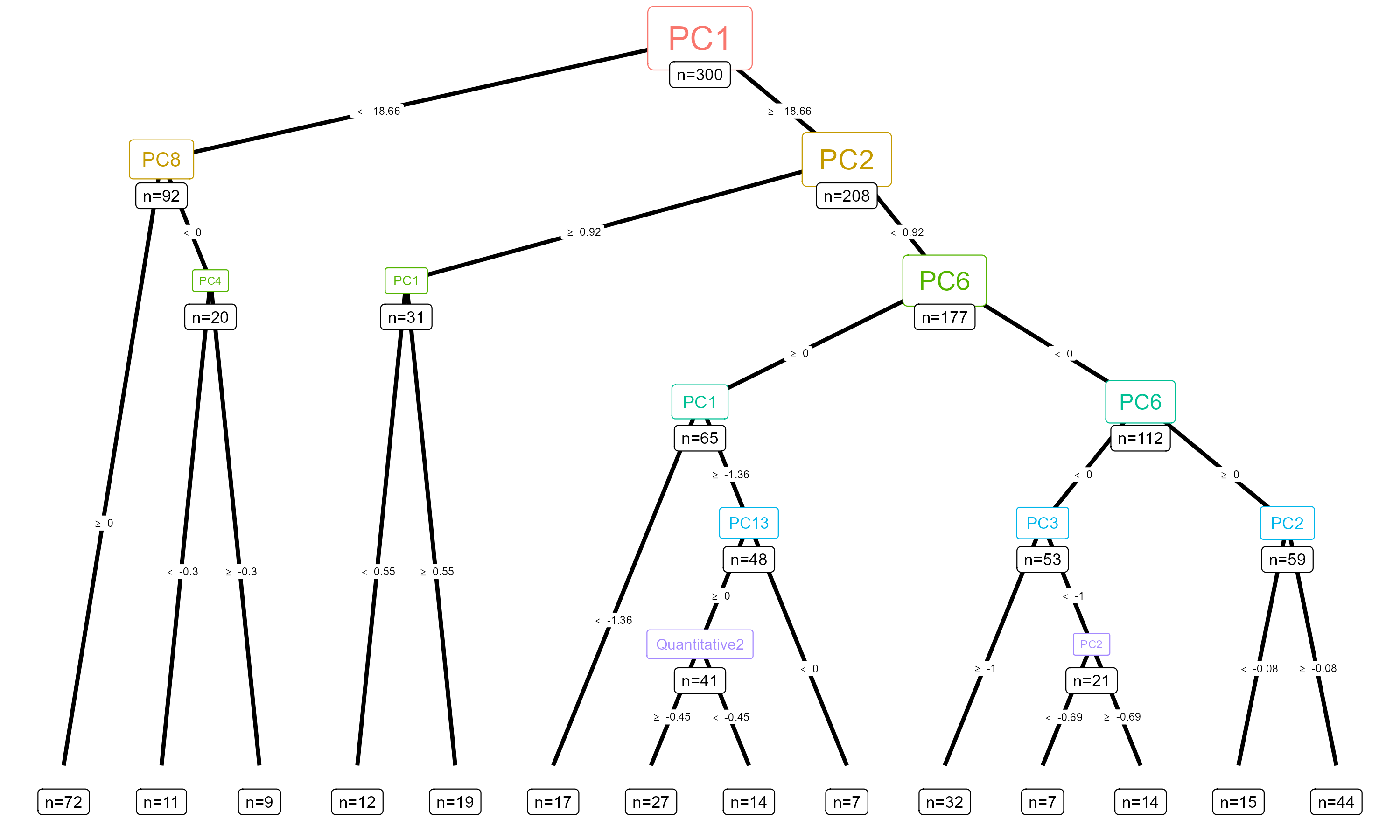} 
    \end{subfigure}
    \hfill
    \begin{subfigure}{0.49\textwidth}
        \centering
        \includegraphics[width=\linewidth]{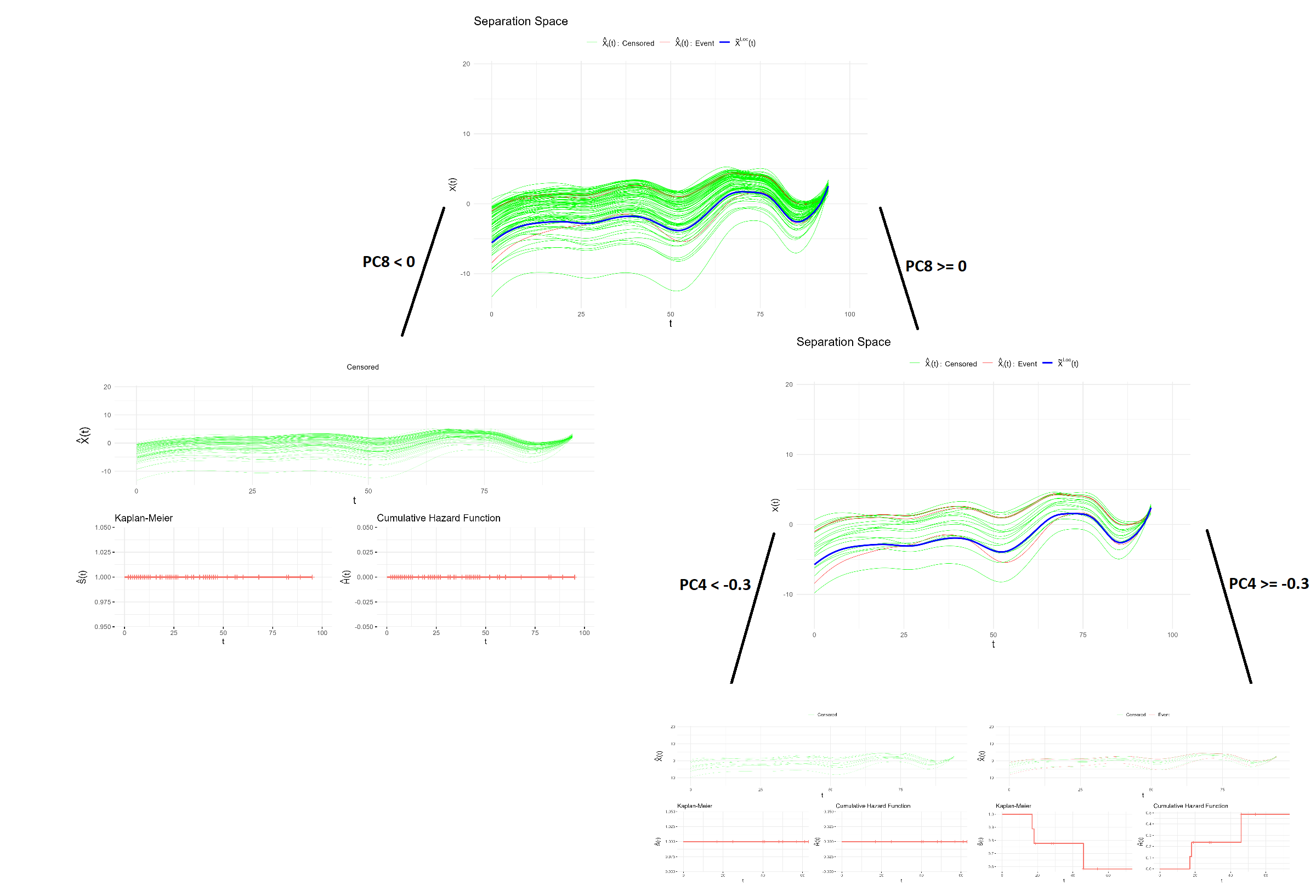} 
    \end{subfigure}
    \caption{Visualization of simulated results for two different sample sizes. The top row corresponds to the case $N=200$ observations: on the left, the FMST, while on the right, a portion of the FMST is displayed in its functional form, highlighting the separation spaces in $\mathcal{L}^2(\mathcal{T})$ norm and the graphical sets at the terminal nodes. The bottom row presents the same layout for a sample $N=300$.}
    \label{fig:tree_simulated}
\end{figure*}

Moreover, the Figure \ref{fig:tree_simulated} illustrates the hierarchical separation process, demonstrating the progressive refinement of survival stratification within an FMST for a selected portion of the structure. The terminal nodes incorporate graphical tools that offer valuable insights into each subject group's condition and characteristics, enhancing the survival analysis's interpretability.

\begin{figure}[h]
    \centering
    \includegraphics[width=0.8\linewidth]{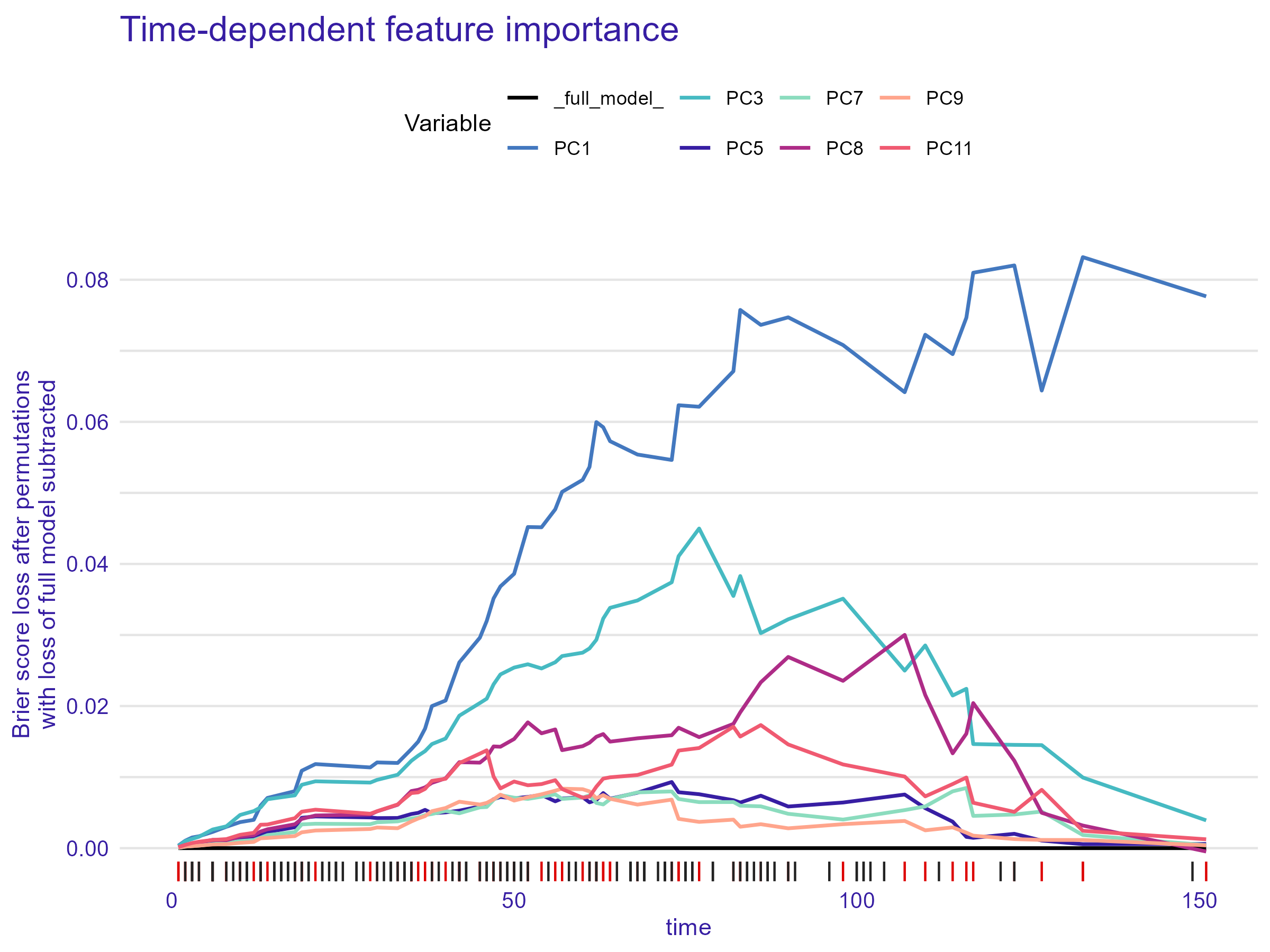}
    \includegraphics[width=0.8\linewidth]{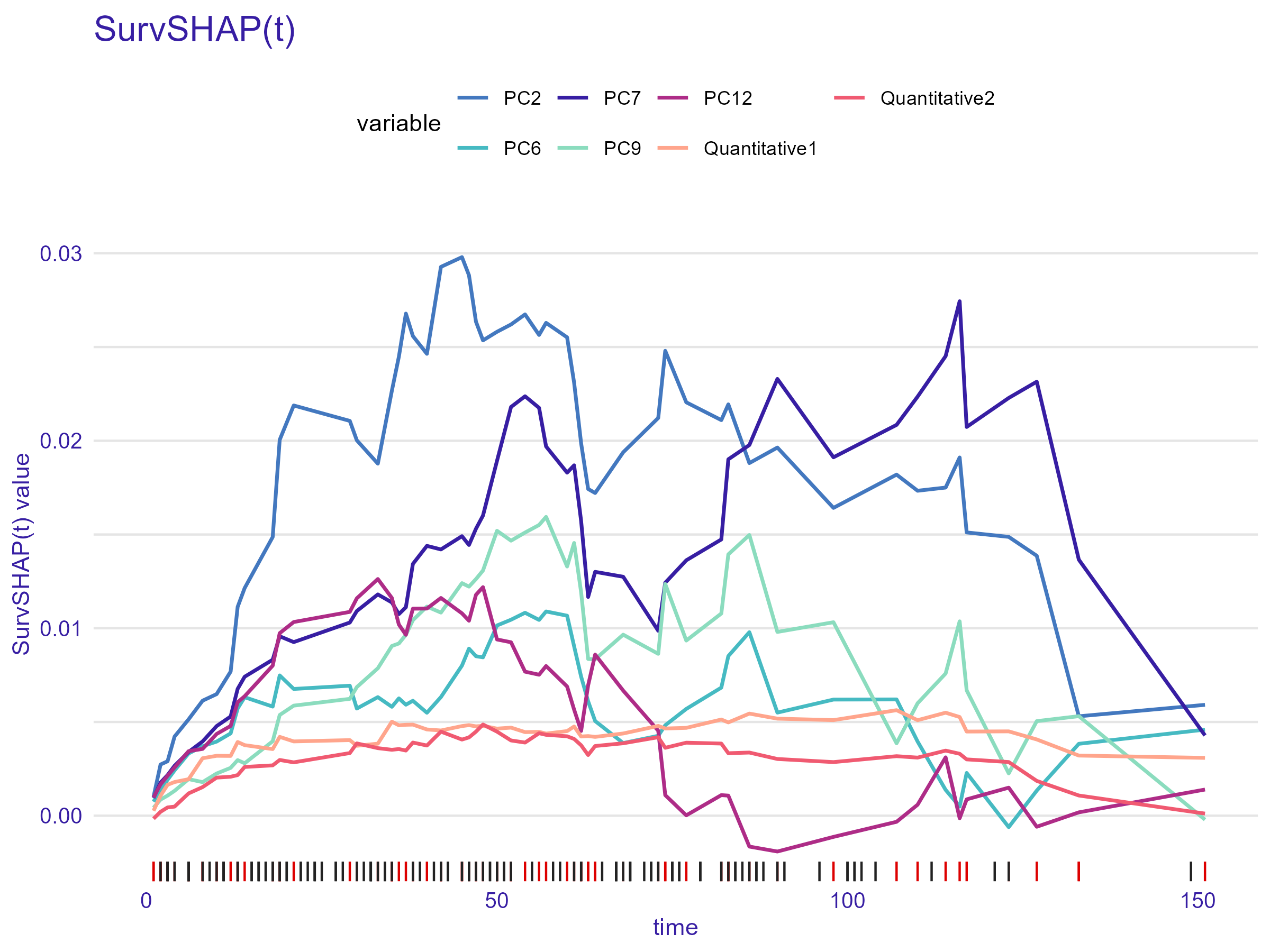}
    \caption{Global (Upper) and local (Lower) explainability representation over time $\mathcal{T}$ for FRSF model.}
    \label{fig:glob_loc_200}
\end{figure}

\begin{figure}[h]
    \centering
    \includegraphics[width=0.8\linewidth]{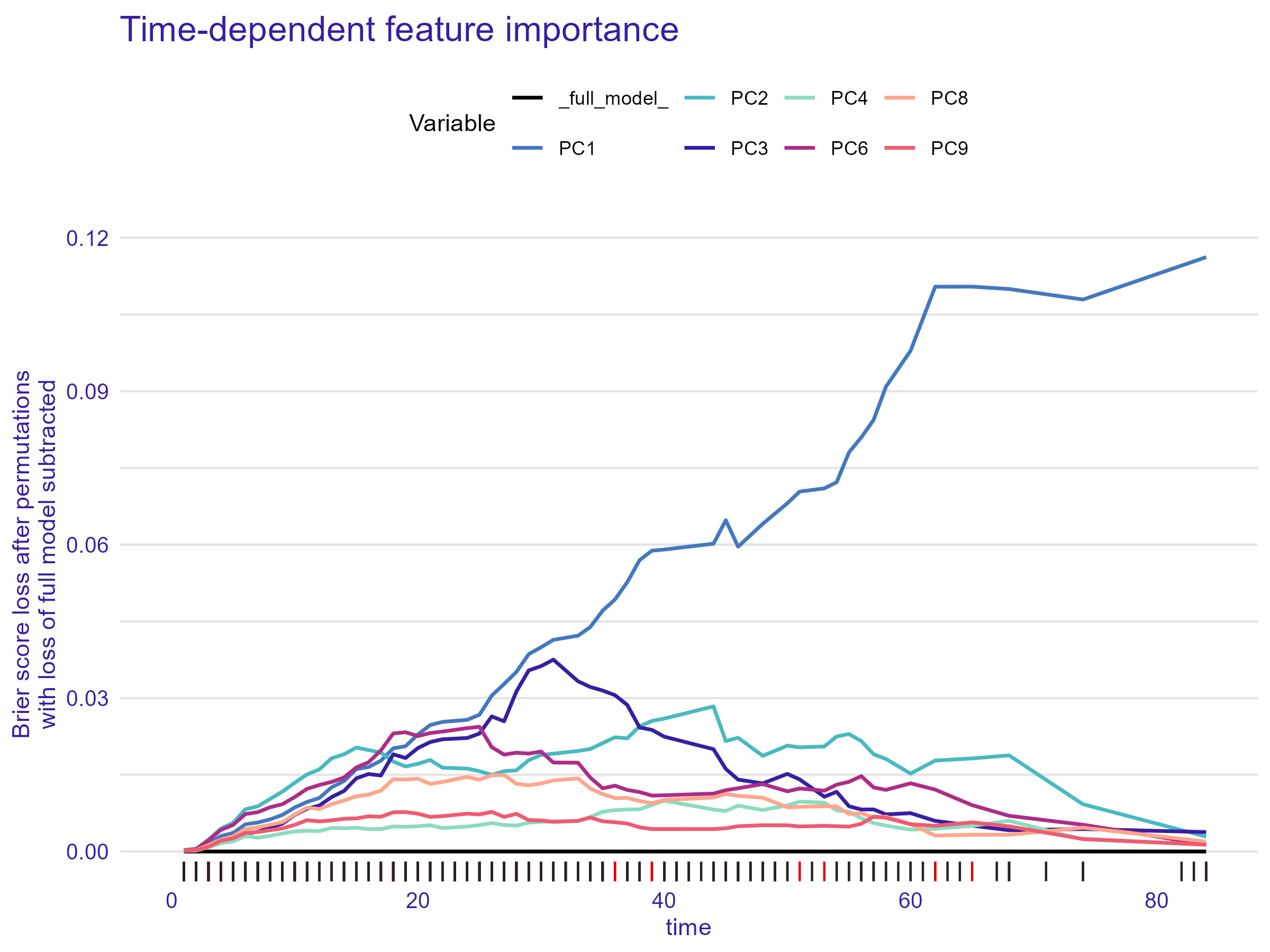}
    \includegraphics[width=0.8\linewidth]{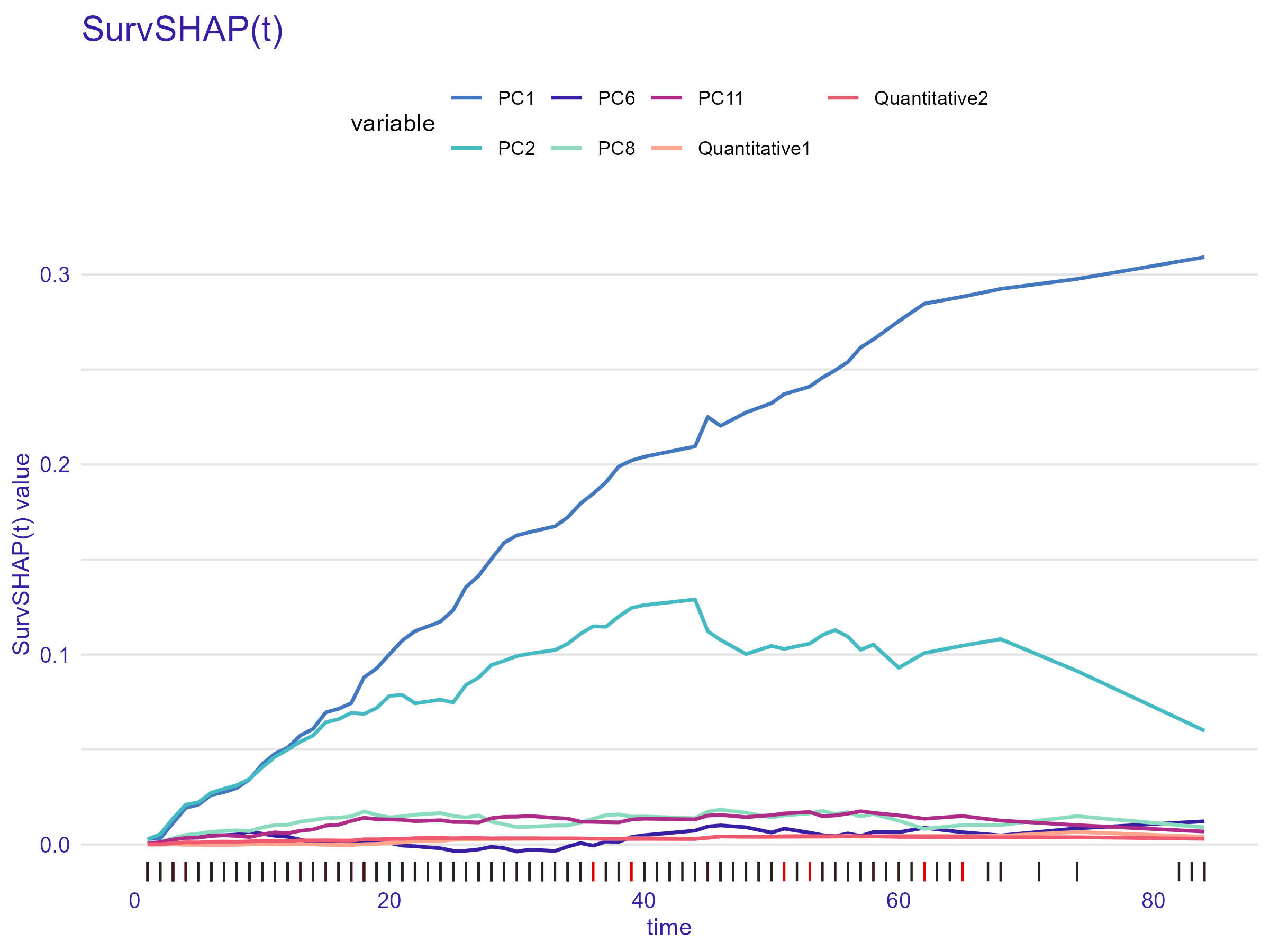}
    \caption{Global (Upper) and local (Lower) explainability representation over time $\mathcal{T}$ for FRSF model.}
    \label{fig:glob_loc_300}
\end{figure}

In the final scenario, we investigated the explainability of FRSF selecting as a number of trees \(B=1000\) and the log-rank test as a splitting method, analyzing the importance of functional covariates at unknown time event points. Given the complexity of survival forests in handling functional data, we extracted variable importance measures across different time intervals, highlighting the temporal relevance of specific covariates using a local and a global approach for the two cases (Figures: \ref{fig:glob_loc_200},\ref{fig:glob_loc_300}). This analysis provided insights into the dynamic contribution of each covariate to survival prediction.  

The results, shown in Appendix (Tables: \ref{table_glob_200},\ref{table_loc_200},\ref{table_glob_300},\ref{table_loc_300}), illustrate how variable importance evolves in the global case and in the local case, emphasizing critical periods where specific covariates substantially influence survival outcomes. This approach enhances the explainability of FRSF models by clearly interpreting how functional covariates impact survival at different instant points.  

In general, these two scenarios collectively provide a comprehensive assessment of the interpretability of functional survival models, demonstrating the effectiveness of the proposed techniques in improving model transparency and understanding.

\subsection{Case study: Sequential Organ Failure Assessment (SOFA)}
In this case study, we examine the application of the Sequential Organ Failure Assessment (SOFA) score within the context of a survival framework that has been developed. The SOFA dataset is in the R package 'refund' \citep{refundR}. A key aspect of this case study is including the time-dependent data, the SOFA score. The SOFA score is used as a dynamic measure of organ dysfunction, reflecting changes over time. Unlike static variables such as age or comorbidities, the SOFA score changes based on the patient's clinical progression, and its time-dependent nature is crucial to monitoring the patient's response to treatment or deterioration of organ function. The SOFA score is calculated periodically in $\mathcal{T}=[1,173]$ where each organ system receives a score based on specific clinical parameters. 

\begin{figure*}[tbp]
    \centering
    \begin{subfigure}{0.49\textwidth}
        \centering
        \includegraphics[width=\linewidth]{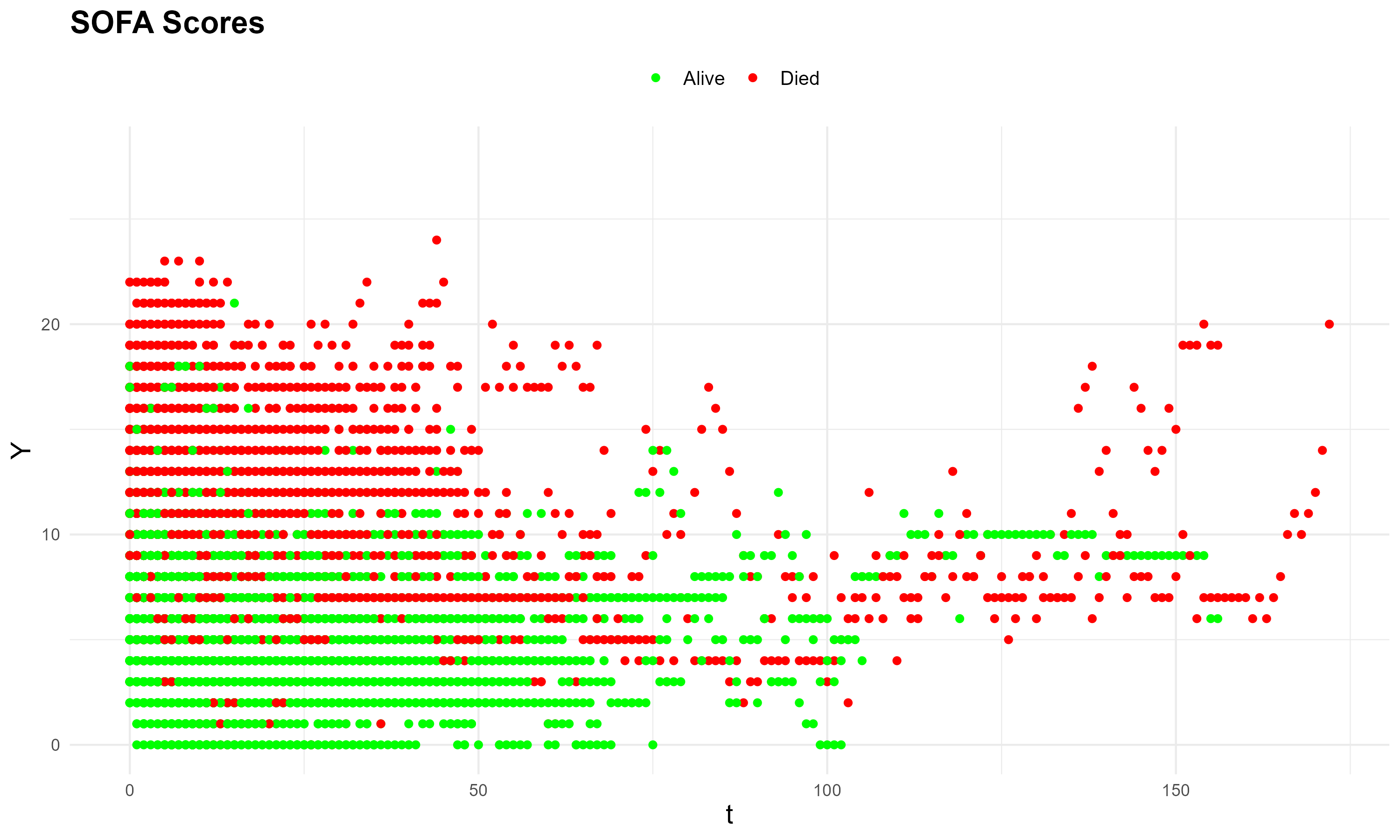} 
    \end{subfigure}
    \hfill
    \begin{subfigure}{0.49\textwidth}
        \centering
        \includegraphics[width=\linewidth]{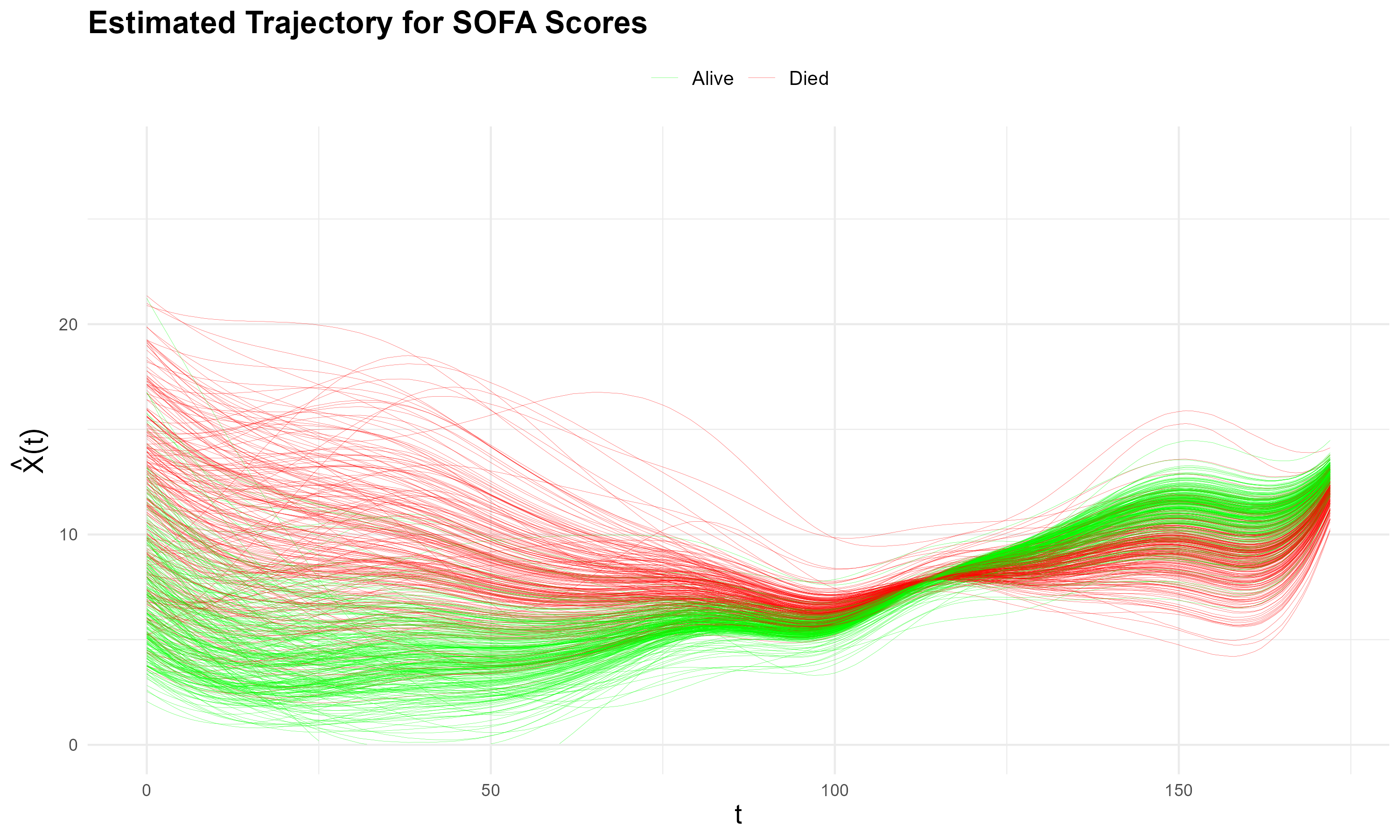} 
    \end{subfigure}
    \caption{Graphical representation of SOFA scores and estimated trajectory over time stratified by outcome.}
    \label{fig:sofa}
\end{figure*}

In this analysis, we aim to model the relationships between the time-dependent effects of the SOFA score alongside other patient characteristics such as age, gender, and comorbidities. From a statistical perspective, the objective is to understand how these variables interact within the functional space, capturing their dynamic influence on patient outcomes. Figure \ref{fig:sofa} illustrates their distribution and temporal evolution obtained by applying the FPCA decomposition by selecting $p=14$ as several components. The functions highlight the variability in patient progression, showing distinct patterns associated with different clinical outcomes.

\begin{figure*}[tbp]
    \centering
    \begin{subfigure}{0.80\linewidth}
    \includegraphics[width=\linewidth]{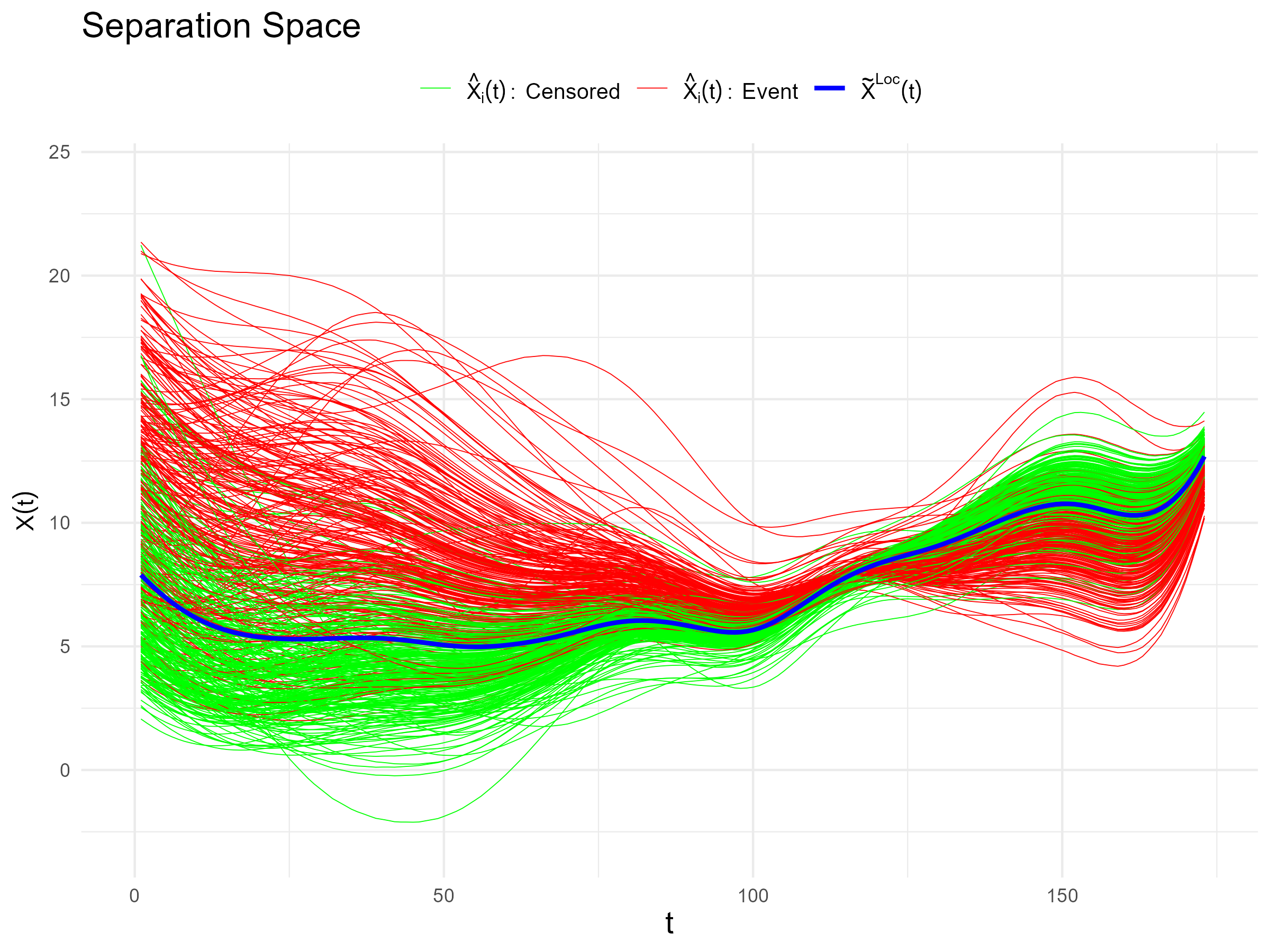}
    \end{subfigure}
    \caption{Functional space separation at the root node of FMST, illustrating the differentiation by LFSDC of patients' trajectories based on survival outcomes.}
    \label{fig:root_sofa}
\end{figure*}

To further analyze the relationship between the variables in the SOFA dataset and clinical outcomes, we applied an FMST. Figure \ref{fig:root_sofa} shows the separation space in the root node highlighting the discriminative capability of LFSDC to stratify patient risk based on functional predictors.

\begin{figure*}[tbp]
    \centering
    \begin{subfigure}{0.95\textwidth}
        \centering
        \captionsetup{font=footnotesize}
        \includegraphics[width=\linewidth]{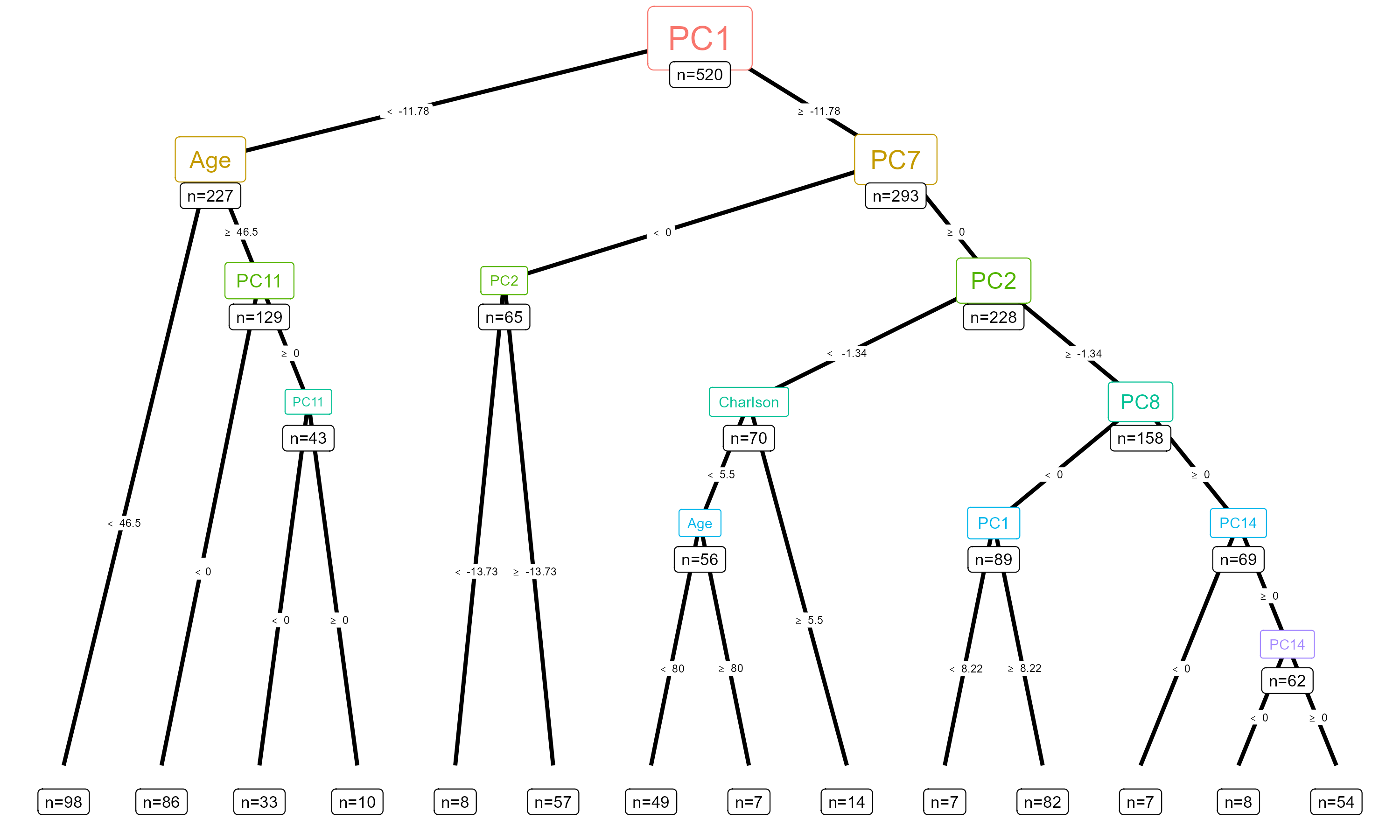} 
        \caption{Functional Mixed Survival Tree (FMST) for SOFA dataset.}
        \label{fig:FST}
    \end{subfigure}
    \vfill
     \begin{subfigure}{0.95\textwidth}
        \centering
        \captionsetup{font=footnotesize}
        \includegraphics[width=\linewidth,height=13cm]{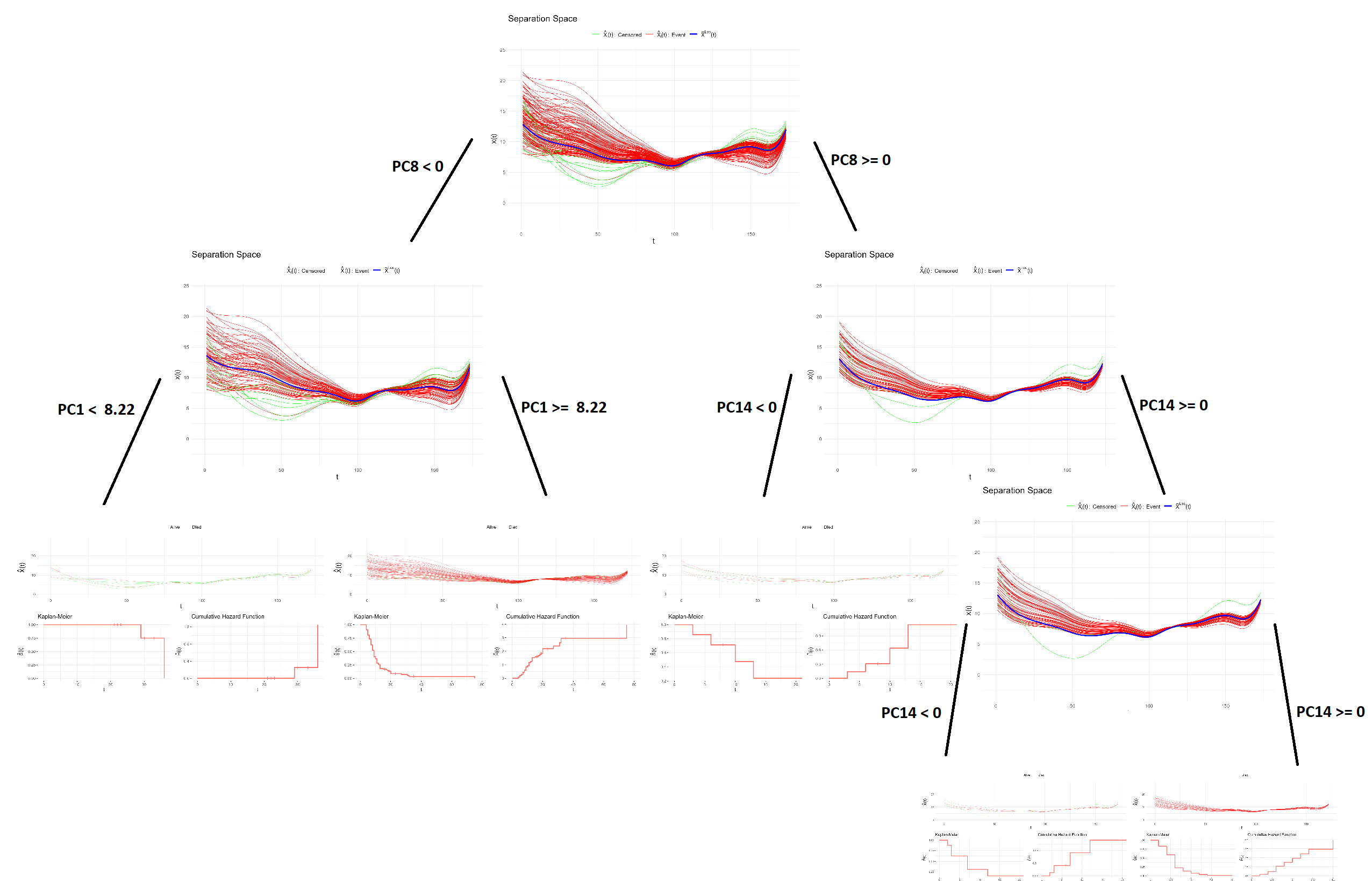} 
        \caption{Graphical representation of the functional space partitioning in the FMST, highlighting the results within terminal nodes.}
        \label{fig:FST_plot}
    \end{subfigure}
    \caption{Graphical results of the FMST, illustrating the tree structure and the partitioning of the functional space at terminal nodes.}
    \label{fig:image_total_1}
\end{figure*}

The resulting tree structure, shown in Figure \ref{fig:FST}, enables the visualization of decision rules based on patient characteristics and SOFA score. We analyse the separation process at each node to better understand the underlying functional process driving separation in the FMST. Since using FPCA components abstracts the original SOFA score trajectories, leading to a loss of direct interpretability regarding clinical progression, we identify the separation space as a means to recover this interpretability. In Figure \ref{fig:FST_plot}, we present a part of the functional graphical representation of the separation process in FMST. Each split node displays the separation space and the corresponding separating curve, highlighting how functional SOFA score trajectories contribute to patient stratification.

\begin{figure*}[tbp]
    \centering
    \begin{subfigure}{0.80\linewidth}
    \includegraphics[width=\linewidth]{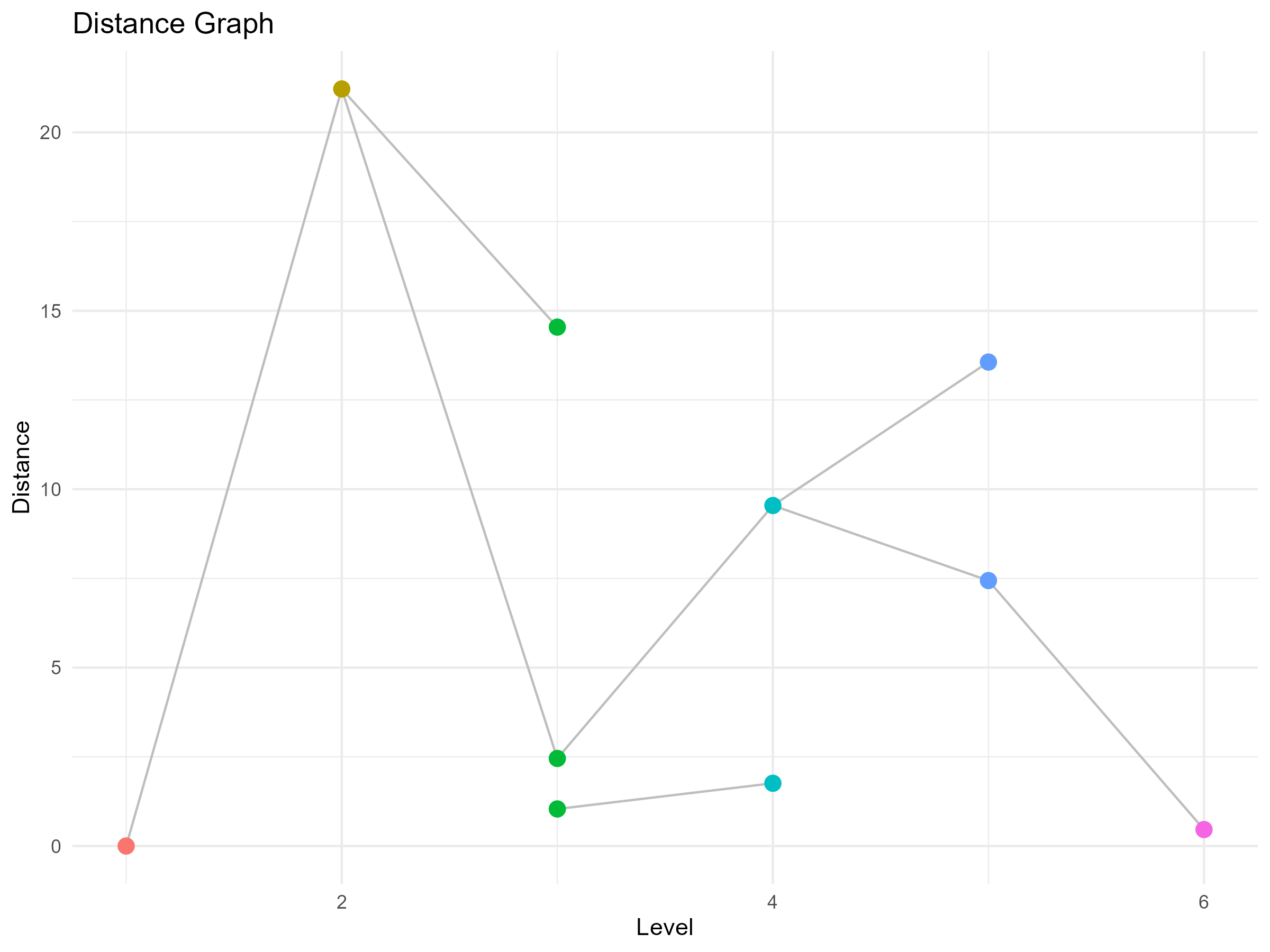}
    \end{subfigure}
    \caption{Survival dynamics of distance at each level for $\mathcal{A}^*_h$ nodes of FMST for dataset SOFA.}
    \label{fig:dist_sofa}
\end{figure*}

Finally, we analyse the evolution of the local separation space to quantify the separation dynamics between the parent node and daughter nodes. Specifically, we compute the distance between the LFSDC curves at each step of the separation process using the metric by Equation \ref{eq:dist_sep}  as shown in Figure \ref{fig:dist_sofa}.

\begin{figure}[h]
    \centering
    \includegraphics[width=0.8\linewidth]{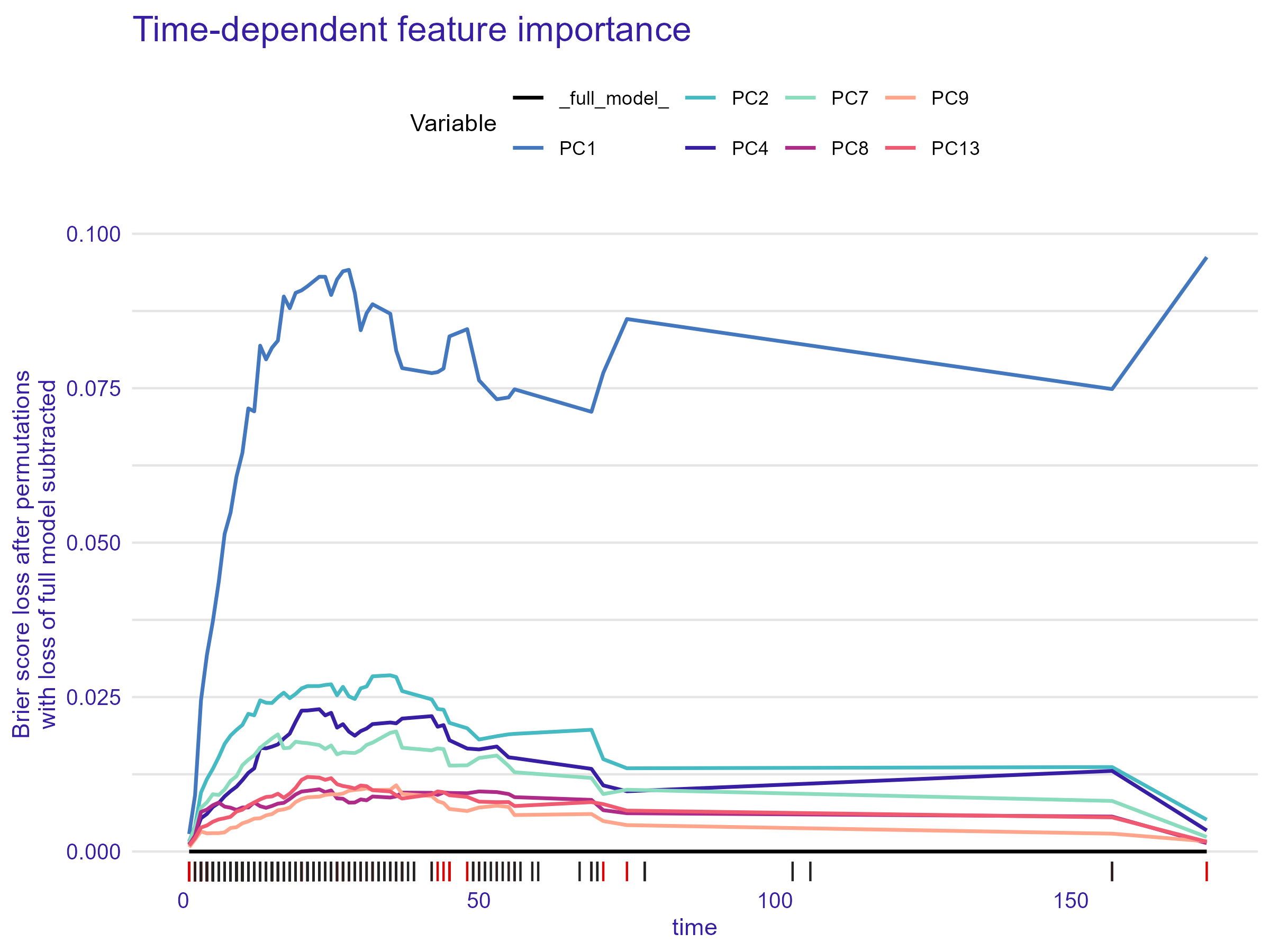}
    \includegraphics[width=0.8\linewidth]{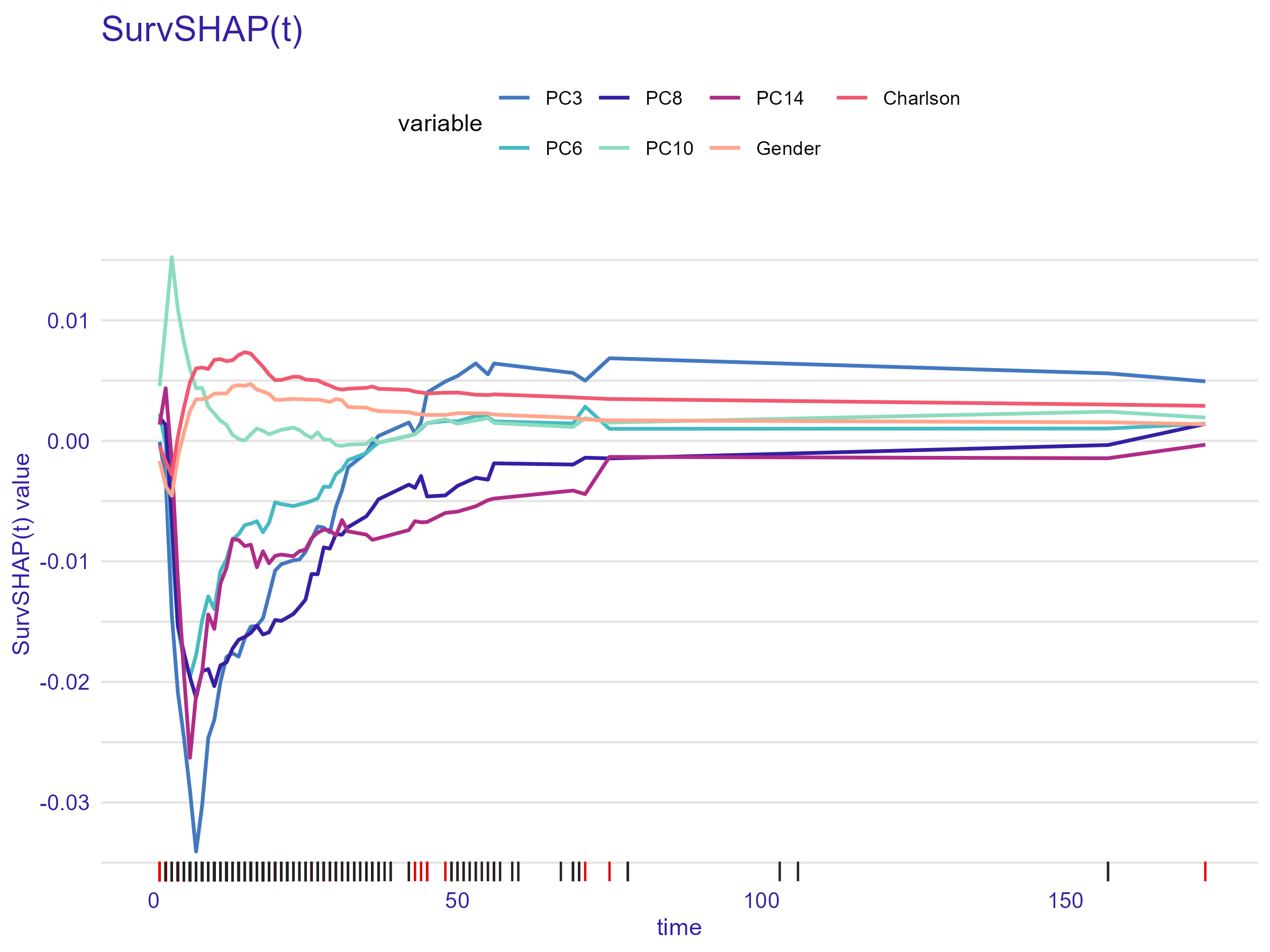}
    \caption{Global (Upper) and local (Lower) explainability representation over time $\mathcal{T}$ for FRSF model.}
    \label{fig:glob_loc_sofa}
\end{figure}

The second step of our analysis involved applying the FRSF, selecting $B=1000$ as the number of STs, to ensure robust and stable predictions. At this point, we focused on global and local explainability to assess variable importance, even when event data are unavailable. By aggregating information from multiple trees, the model smooths survival probabilities at unobserved time points, providing a more comprehensive view of patient risk evolution. Table \ref{fig:glob_loc_sofa} summarises the variable importance scores derived from the FRSF, highlighting the contributions of baseline covariates and functional SOFA score components to survival predictions. 

In particular, the PC2 component of the SOFA score, analysed through global explainability (Table \ref{table_glob_sofa}), reveals relatively stable but low importance scores over time, with few fluctuations. However, there are critical intervals such as $[56,69]$ and $[75,157]$ where a marked increase in the cumulative contribution is observed, as indicated by high $MTGD$ and $MTNGD$ values. These segments suggest moments where this feature becomes more influential in shaping survival probabilities, possibly reflecting phases of acute patient deterioration or recovery.

On the other hand, the local explainability analysis for PC10 in the second unit (Table \ref{table_loc_sofa}) highlights how individual contributions can deviate from global trends. While most intervals show marginal influence, the time window $[75,157]$ exhibits a notably large negative $TSD$ value, suggesting a strong individualised effect that diverges from the general behaviour of the variable. This discrepancy underscores the added value of local explanations in detecting subject-specific dynamics that may not emerge in aggregated analyses.

The graphical comparisons (Figures: \ref{recost_sofa_global},\ref{recost_loc_sofa}) further emphasize the advantage of the functional reconstruction approach over discrete approximations, offering smoother and more interpretable insights into the time-evolving relevance of covariates. Notably, regions with no observed events—marked by dashed vertical lines—highlight the robustness of the functional smoothing strategy, which still extracts meaningful trends in feature influence despite the lack of event data.

Overall, the proposed framework allows a richer understanding of how each functional or scalar covariate interacts with survival dynamics, promoting global explainability and personalised clinical insights.

\section{Discussion and Conclusions}

This study aims to tackle a significant challenge in using functional survival models: the limited interpretability of Functional Survival Trees (FST) and the complexity involved in explaining predictions from ensemble methods, such as the Functional Random Survival Forest (FRSF). Our goal was to bridge this gap by introducing methodological and visual tools that provide both local and global insights into the behavior of these models, particularly in the context of time-to-event data with functional and mixed covariates.
We introduced two main contributions. First, for FST models, we developed a graphical interpretability framework based on the Local Functional Survival Discrimination Curve (LFSDC) and a node-level separability metric. These tools enable a clear, stepwise understanding of how survival heterogeneity evolves along the tree structure, allowing practitioners to visualize how specific functional components influence survival probabilities at different locations. This level of transparency is particularly valuable in medical settings, where decision paths must be interpretable and justifiable.
Second, we addressed the explainability of FRSF by proposing a dual-level approach that includes both local and global perspectives. At the local level, we adapted the Shapley Additive Explanation (SHAP) methodology to survival analysis with functional and mixed predictors through the SurvSHAP$(t)$ algorithm, quantifying how classical and functional individual features contribute to survival predictions over time. At the global level, we implemented Permutation Feature Importance (PFI), adapting it to handle time-dependent relevance and heterogeneity in functional covariates. Together, these methods comprehensively explain how feature information is used across the entire ensemble.

Our simulation studies confirmed that the proposed tools faithfully retrieve the internal decision mechanisms of survival trees and accurately highlight the time-varying importance of functional components. In particular, we showed that the proposed graphical tools effectively capture separation processes within functional trees, and that local SurvSHAP curves successfully trace the temporal dynamics of variable contributions. The case study on SOFA scores in intensive care patients further illustrated our methods' practical relevance, showing how functional predictors and baseline characteristics can be jointly analysed to yield clinically interpretable results.

From a positioning standpoint, our work contributes to the growing field of FDA and explainable machine learning in healthcare by extending interpretability and explainability techniques to functional survival analysis, an area where high model complexity has often hindered practical usability. By situating our methods within the broader goals of explainable AI (XAI), we aim to promote predictive accuracy, trust, and transparency in high-stakes applications.

Nevertheless, our study may have some limitations. First, the computational cost of local explainability tools such as SurvSHAP can be significant, especially when working with high-dimensional or irregular time series. Then, the generalizability of visual tools depends on the quality of the estimated functional trajectories, which in turn are sensitive to data sparsity and noise.
Future work will extend these methods to alternative functional representations (e.g., wavelet bases), improve computational efficiency through approximated SHAP variants, and validate the proposed framework across larger and more diverse real-world datasets. Moreover, we plan to investigate how these interpretability tools can be integrated into dynamic clinical decision support systems, where accuracy and understandability are essential.
In conclusion, this study provides a unified framework for interpreting functional survival models, combining rigorous statistical methodology with visual and explainability tools. 




\bibliographystyle{abbrvnat}
\bibliography{wileyNJD-AMA.bib}%

\clearpage

\onecolumn

\section*{Supplemental Material}
\addcontentsline{toc}{section}{Supplemental Material}

\subsection*{Simulation study for $N=200$}

\begin{table}[h]
\Large
\centering
\begin{minipage}{0.54\textwidth}
\centering
\renewcommand{\arraystretch}{1.2} 
\setlength{\tabcolsep}{19pt} 
\resizebox{\textwidth}{!}{ 
\begin{tabular}{c|cc|c|c} 
\hline
\hline
\multicolumn{5}{c}{\textbf{Global Explainability for PC1}} \\  
\hline
\hline
\multirow{2}{*}{\([t_\alpha,t_\beta]\)} & \multicolumn{2}{c|}{\(\overline{FI}_1(t)\)} & \multirow{2}{*}{\(MTGD_{t_\alpha,t_\beta}^{1}\)} &\multirow{2}{*} {\(MTNGD_{t_\alpha,t_\beta}^{1}\)} \\ \cline{2-3}
                        & \(\overline{FI}_1(t_\alpha)\) & \(\overline{FI}_1(t_\beta)\) &  &  \\ 
\hline
[4,6]                    & 0.00185283 & 0.00245758 & $-5.38 \times 10^{-6}$ & $-2.69 \times 10^{-6}$ \\ \hline
[6,8]                    & 0.00245758 & 0.00316047 & $1.49 \times 10^{-5}$  & $7.45 \times 10^{-6}$  \\ \hline
[8,10]                   & 0.00316047 & 0.00355074 & $6.46 \times 10^{-5}$  & $3.23 \times 10^{-5}$  \\ \hline
[10,12]                  & 0.00355074 & 0.00394254 & $-3.04 \times 10^{-4}$ & $-1.52 \times 10^{-4}$ \\ \hline
[14,18]                  & 0.00679740 & 0.00730410 & $-1.65 \times 10^{-3}$ & $-4.12 \times 10^{-4}$ \\ \hline
[19,21]                  & 0.00991108 & 0.01144837 & $4.16 \times 10^{-4}$  & $2.08 \times 10^{-4}$  \\ \hline
[21,29]                  & 0.01144837 & 0.01098038 & $3.05 \times 10^{-3}$  & $3.82 \times 10^{-4}$  \\ \hline
[30,33]                  & 0.01195368 & 0.01174326 & $-1.03 \times 10^{-3}$ & $-3.43 \times 10^{-4}$ \\ \hline
[33,35]                  & 0.01174326 & 0.01404944 & $-2.22 \times 10^{-4}$ & $-1.11 \times 10^{-4}$ \\ \hline
[38,40]                  & 0.02025410 & 0.02124029 & $-1.14 \times 10^{-4}$ & $-5.70 \times 10^{-5}$ \\ \hline
[40,42]                  & 0.02124029 & 0.02717545 & $-1.94 \times 10^{-4}$ & $-9.69 \times 10^{-5}$ \\ \hline
[42,45]                  & 0.02717545 & 0.03056935 & $7.71 \times 10^{-4}$  & $2.57 \times 10^{-4}$  \\ \hline
[48,50]                  & 0.03657040 & 0.03724537 & $-1.91 \times 10^{-4}$ & $-9.53 \times 10^{-5}$ \\ \hline
[50,52]                  & 0.03724537 & 0.04120939 & $4.77 \times 10^{-4}$  & $2.38 \times 10^{-4}$  \\ \hline
[52,54]                  & 0.04120939 & 0.04068467 & $-3.00 \times 10^{-4}$ & $-1.50 \times 10^{-4}$ \\ \hline
[54,56]                  & 0.04068467 & 0.04340819 & $-3.17 \times 10^{-4}$ & $-1.58 \times 10^{-4}$ \\ \hline
[57,60]                  & 0.04641078 & 0.04826712 & $-2.23 \times 10^{-4}$ & $-7.42 \times 10^{-5}$ \\ \hline
[64,68]                  & 0.05685593 & 0.05357206 & $2.83 \times 10^{-4}$  & $7.09 \times 10^{-5}$  \\ \hline
[68,73]                  & 0.05357206 & 0.05410605 & $-1.12 \times 10^{-2}$ & $-2.24 \times 10^{-3}$ \\ \hline
[74,77]                  & 0.06068427 & 0.06309772 & $3.53 \times 10^{-3}$  & $1.18 \times 10^{-3}$  \\ \hline
[77,82]                  & 0.06309772 & 0.06709227 & $-1.42 \times 10^{-2}$ & $-2.84 \times 10^{-3}$ \\ \hline
[83,86]                  & 0.07438620 & 0.07298383 & $4.72 \times 10^{-3}$  & $1.57 \times 10^{-3}$  \\ \hline
[86,90]                  & 0.07298383 & 0.07278383 & $-2.83 \times 10^{-3}$ & $-7.07 \times 10^{-4}$ \\ \hline
[90,98]                  & 0.07278383 & 0.06751300 & $1.09 \times 10^{-2}$  & $1.37 \times 10^{-3}$  \\ \hline
[98,107]                 & 0.06751300 & 0.06186418 & $-2.58 \times 10^{-2}$ & $-2.87 \times 10^{-3}$ \\ \hline
[107,110]                & 0.06186418 & 0.06943421 & $6.17 \times 10^{-4}$  & $2.06 \times 10^{-4}$  \\ \hline
[110,114]                & 0.06943421 & 0.06965263 & $5.69 \times 10^{-4}$  & $1.42 \times 10^{-4}$  \\ \hline
[114,116]                & 0.06965263 & 0.07486795 & $-1.20 \times 10^{-3}$ & $-6.01 \times 10^{-4}$ \\ \hline
[117,123]                & 0.08150482 & 0.07812225 & $2.70 \times 10^{-2}$  & $4.51 \times 10^{-3}$  \\ \hline
[123,127]                & 0.07812225 & 0.06493811 & $-2.77 \times 10^{-3}$ & $-6.94 \times 10^{-4}$ \\ \hline
[127,133]                & 0.06493811 & 0.07349230 & $-1.54 \times 10^{-2}$ & $-2.57 \times 10^{-3}$ \\ \hline
[133,151]                & 0.07349230 & 0.07439249 & $3.94 \times 10^{-1}$  & $2.19 \times 10^{-2}$  \\ \hline \hline
\end{tabular}
}
\caption{Numerical results in PFI for PC1.}
\label{table_glob_200}
\end{minipage} \vfill
\vspace{5mm}
\begin{minipage}{0.68\textwidth}
\centering
\includegraphics[width=\linewidth]{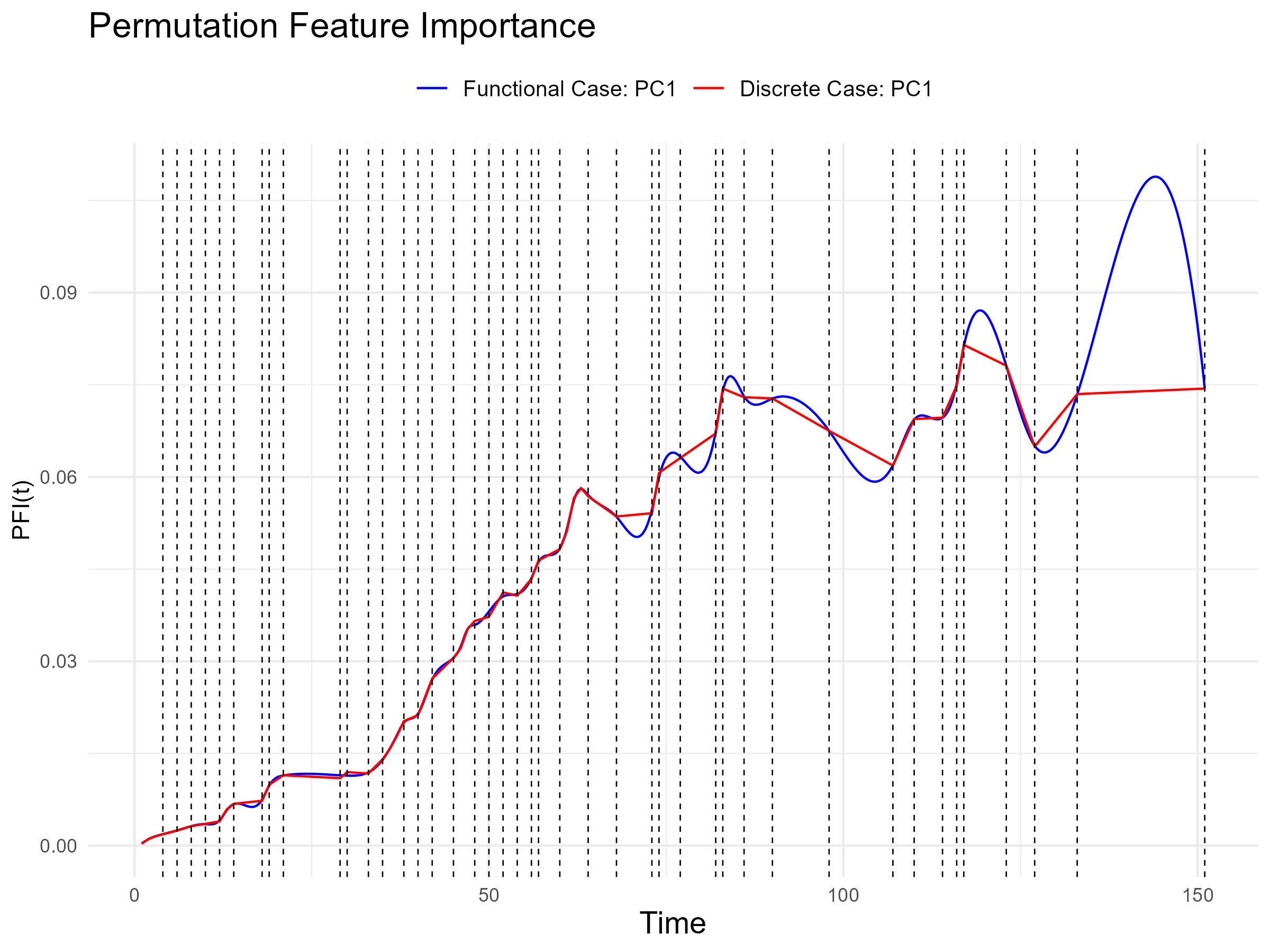}
\captionof{figure}{Global explainability of PC1 in FRSF using PFI. The discrete case (red line) is compared to the functional reconstruction (blue line). Vertical black dashed lines indicate time intervals with no observed events.}
\label{fig:your_image1}
\end{minipage}
\end{table}  

\begin{table}[h]
\Large
\centering
\begin{minipage}{0.54\textwidth}
\centering
\renewcommand{\arraystretch}{1.2} 
\setlength{\tabcolsep}{19pt} 
\resizebox{\textwidth}{!}{ 
\begin{tabular}{c|cc|c|c} 
\hline
\hline
\multicolumn{5}{c}{\textbf{Local Explainability for PC2 for $2$th unit}} \\  
\hline
\hline
\multirow{2}{*}{\([t_\alpha,t_\beta]\)} & \multicolumn{2}{c|}{\(\phi^*_t(\bm{w_*},2)\)} & \multirow{2}{*}{\(TSD_{t_\alpha,t_\beta}^{22}\)} &\multirow{2}{*} {\(TNSD_{t_\alpha,t_\beta}^{22}\)} \\ \cline{2-3}
                        & \(\phi^*_{t_\alpha}(\bm{w_*},2)\) & \(\phi^*_{t_\beta}(\bm{w_*},2)\) &  &  \\ 
\hline
[4,6]                    & 0.004221 & 0.005140   & $2.56 \times 10^{-5}$  & $7.82 \times 10^{-6}$  \\ \hline
[6,8]                    & 0.005140 & 0.006136   & $3.69 \times 10^{-5}$  & $1.84 \times 10^{-5}$  \\ \hline
[8,10]                   & 0.006136 & 0.006486   & $-1.92 \times 10^{-4}$ & $-9.60 \times 10^{-5}$ \\ \hline
[10,12]                  & 0.006486 & 0.007689   & $1.52 \times 10^{-4}$  & $7.61 \times 10^{-5}$  \\ \hline
[14,18]                  & 0.012143 & 0.014865   & $-3.68 \times 10^{-4}$ & $-9.20 \times 10^{-5}$ \\ \hline
[19,21]                  & 0.020052 & 0.021882   & $-2.66 \times 10^{-4}$ & $-1.33 \times 10^{-4}$ \\ \hline
[21,29]                  & 0.021882 & 0.021060   & $9.31 \times 10^{-3}$  & $1.16 \times 10^{-3}$  \\ \hline
[30,33]                  & 0.020025 & 0.018780   & $-8.92 \times 10^{-4}$ & $-2.97 \times 10^{-4}$ \\ \hline
[33,35]                  & 0.018780 & 0.022667   & $-1.86 \times 10^{-4}$ & $-9.31 \times 10^{-5}$ \\ \hline
[38,40]                  & 0.025583 & 0.024638   & $3.55 \times 10^{-4}$  & $1.78 \times 10^{-4}$  \\ \hline
[40,42]                  & 0.024638 & 0.029275   & $3.62 \times 10^{-5}$  & $1.81 \times 10^{-5}$  \\ \hline
[42,45]                  & 0.029275 & 0.029794   & $1.37 \times 10^{-3}$  & $4.56 \times 10^{-4}$  \\ \hline
[48,50]                  & 0.025354 & 0.025813   & $-2.62 \times 10^{-4}$ & $-1.31 \times 10^{-4}$ \\ \hline
[50,52]                  & 0.025813 & 0.026203   & $-1.76 \times 10^{-4}$ & $-8.79 \times 10^{-5}$ \\ \hline
[52,54]                  & 0.026204 & 0.026740   & $1.21 \times 10^{-4}$  & $7.46 \times 10^{-5}$  \\ \hline
[54,56]                  & 0.026740 & 0.025645   & $-5.77 \times 10^{-7}$ & $-2.89 \times 10^{-7}$ \\ \hline
[57,60]                  & 0.026289 & 0.025516   & $1.01 \times 10^{-3}$  & $3.37 \times 10^{-4}$  \\ \hline
[64,68]                  & 0.017213 & 0.019385   & $-1.75 \times 10^{-3}$ & $-4.36 \times 10^{-4}$ \\ \hline
[68,73]                  & 0.019385 & 0.021216   & $2.19 \times 10^{-3}$  & $4.39 \times 10^{-4}$  \\ \hline
[74,77]                  & 0.024798 & 0.022050   & $-3.46 \times 10^{-4}$ & $-1.15 \times 10^{-4}$ \\ \hline
[77,82]                  & 0.022050 & 0.021102   & $-1.47 \times 10^{-3}$ & $-2.94 \times 10^{-4}$ \\ \hline
[83,86]                  & 0.021942 & 0.018810   & $9.79 \times 10^{-4}$  & $3.26 \times 10^{-4}$  \\ \hline
[86,90]                  & 0.018810 & 0.019638   & $-1.07 \times 10^{-3}$ & $-2.67 \times 10^{-4}$ \\ \hline
[90,98]                  & 0.019638 & 0.016420   & $2.50 \times 10^{-3}$  & $3.12 \times 10^{-4}$  \\ \hline
[98,107]                 & 0.016420 & 0.018195   & $-1.03 \times 10^{-3}$ & $-1.15 \times 10^{-4}$ \\ \hline
[107,110]                & 0.018195 & 0.017333   & $1.21 \times 10^{-4}$  & $4.04 \times 10^{-5}$  \\ \hline
[110,114]                & 0.017333 & 0.017504   & $-2.77 \times 10^{-4}$ & $-6.93 \times 10^{-5}$ \\ \hline
[114,116]                & 0.017504 & 0.019109   & $-2.61 \times 10^{-4}$ & $-1.30 \times 10^{-4}$ \\ \hline
[117,123]                & 0.015110 & 0.014873   & $-3.72 \times 10^{-3}$ & $-6.21 \times 10^{-4}$ \\ \hline
[123,127]                & 0.014873 & 0.013865   & $1.29 \times 10^{-3}$  & $3.22 \times 10^{-4}$  \\ \hline
[127,133]                & 0.013865 & 0.005303   & $2.24 \times 10^{-3}$  & $3.73 \times 10^{-4}$  \\ \hline
[133,151]                & 0.005303 & 0.005914   & $-1.44 \times 10^{-1}$ & $-8.01 \times 10^{-3}$ \\ \hline\hline
\end{tabular}
}
\caption{Numerical results of SurvSHAP for PC2 on the 2nd unit.}
\label{table_loc_200}
\end{minipage} \vfill
\vspace{5mm}
\begin{minipage}{0.68\textwidth}
\centering
\includegraphics[width=\linewidth]{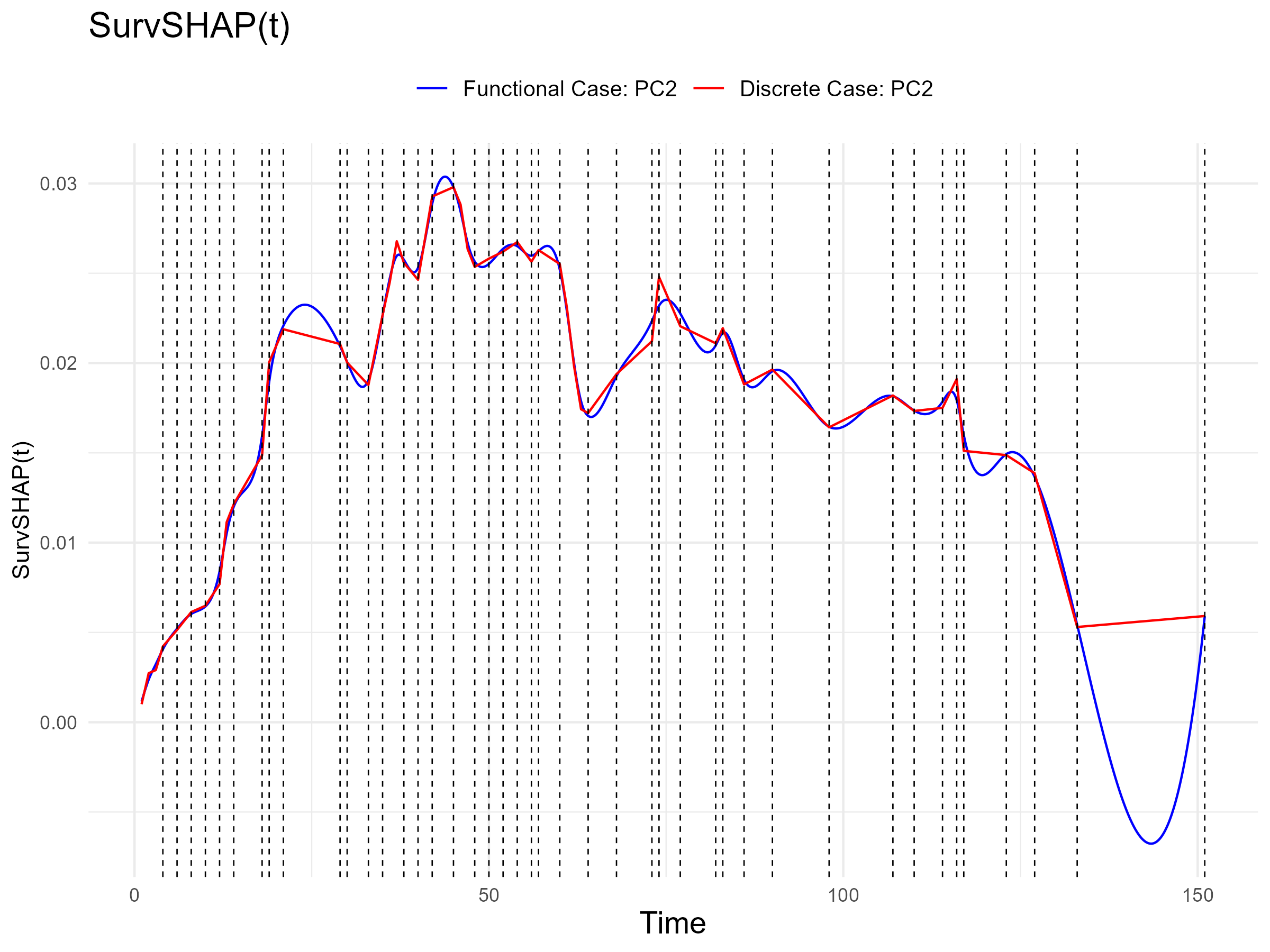}
\captionof{figure}{Local explainability of PC2 in FRSF for 2nd unit. The discrete case (red line) is compared to the functional reconstruction (blue line). Vertical black dashed lines indicate time intervals with no observed events.}
\label{fig:your_image2}
\end{minipage}
\end{table}

\clearpage

\subsection*{Simulation study for $N=300$}

\begin{table}[h]
\Large
\centering
\begin{minipage}{0.70\textwidth}
\centering
\renewcommand{\arraystretch}{1.2} 
\setlength{\tabcolsep}{19pt} 
\resizebox{\textwidth}{!}{ 
\begin{tabular}{c|cc|c|c} 
\hline
\hline
\multicolumn{5}{c}{\textbf{Global Explainability for PC1}} \\  
\hline
\hline
\multirow{2}{*}{\([t_\alpha,t_\beta]\)} & \multicolumn{2}{c|}{\(\overline{FI}_1(t)\)} & \multirow{2}{*}{\(MTGD_{t_\alpha,t_\beta}^{1}\)} &\multirow{2}{*} {\(MTNGD_{t_\alpha,t_\beta}^{1}\)} \\ \cline{2-3}
                        & \(\overline{FI}_1(t_\alpha)\) & \(\overline{FI}_1(t_\beta)\) &  &  \\ 
\hline
[22,24]                  & 0.025342 & 0.025739 & $7.02 \times 10^{-6}$   & $3.51 \times 10^{-6}$   \\ \hline
[31,33]                  & 0.041386 & 0.042205 & $-1.79 \times 10^{-4}$  & $-8.96 \times 10^{-5}$  \\ \hline
[40,44]                  & 0.059048 & 0.060188 & $-7.20 \times 10^{-3}$  & $-1.80 \times 10^{-3}$  \\ \hline
[46,48]                  & 0.059602 & 0.064075 & $-2.38 \times 10^{-3}$  & $-1.19 \times 10^{-3}$  \\ \hline
[48,50]                  & 0.064075 & 0.068107 & $1.02 \times 10^{-3}$   & $5.08 \times 10^{-4}$   \\ \hline
[51,53]                  & 0.070386 & 0.071006 & $1.70 \times 10^{-4}$   & $8.50 \times 10^{-5}$   \\ \hline
[58,60]                  & 0.090880 & 0.097922 & $1.21 \times 10^{-4}$   & $6.07 \times 10^{-5}$   \\ \hline
[60,62]                  & 0.097922 & 0.110446 & $2.09 \times 10^{-4}$   & $1.04 \times 10^{-4}$   \\ \hline
[62,65]                  & 0.110446 & 0.110457 & $4.14 \times 10^{-3}$   & $1.38 \times 10^{-3}$   \\ \hline
[65,68]                  & 0.110457 & 0.109992 & $-1.01 \times 10^{-3}$  & $-3.37 \times 10^{-4}$  \\ \hline
[68,74]                  & 0.109992 & 0.107939 & $1.89 \times 10^{-3}$   & $3.15 \times 10^{-4}$   \\ \hline
[74,84]                  & 0.107939 & 0.116225 & $-2.99 \times 10^{-2}$  & $-2.99 \times 10^{-3}$  \\ \hline \hline
\end{tabular}
}
\caption{Numerical results in PFI for PC1.}
\label{table_glob_300}
\end{minipage} \vfill
\vspace{5mm}
\begin{minipage}{0.68\textwidth}
\centering
\includegraphics[width=\linewidth]{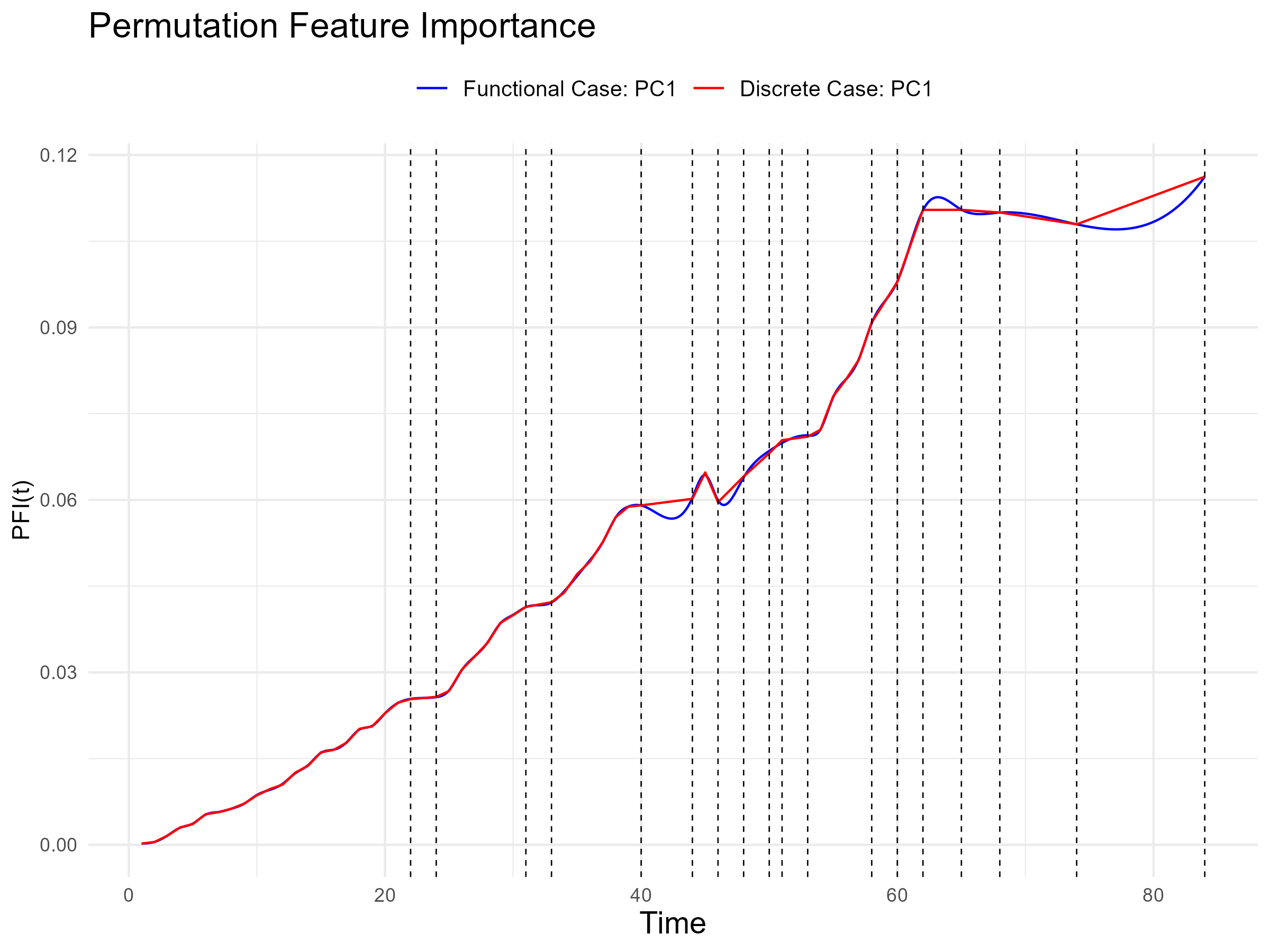}
\captionof{figure}{PFI-based global explainability of PC1 in FRSF, comparing the discrete case (red) and the functional reconstruction (blue). Black dashed lines mark intervals with no observed events.}
\label{fig:your_image3}
\end{minipage}
\end{table}  

\begin{table}[h]
\Large
\centering
\begin{minipage}{0.70\textwidth}
\centering
\renewcommand{\arraystretch}{1.2} 
\setlength{\tabcolsep}{19pt} 
\resizebox{\textwidth}{!}{ 
\begin{tabular}{c|cc|c|c} 
\hline
\hline
\multicolumn{5}{c}{\textbf{Local Explainability for PC2 for $3$th unit}} \\  
\hline
\hline
\multirow{2}{*}{\([t_\alpha,t_\beta]\)} & \multicolumn{2}{c|}{\(\phi^*_t(\bm{w_*},2)\)} & \multirow{2}{*}{\(TSD_{t_\alpha,t_\beta}^{32}\)} &\multirow{2}{*} {\(TNSD_{t_\alpha,t_\beta}^{32}\)} \\ \cline{2-3}
                        & \(\phi^*_{t_\alpha}(\bm{w_*},2)\) & \(\phi^*_{t_\beta}(\bm{w_*},2)\) &  &  \\ 
\hline
[22,24]                    & 0.074305 & 0.076221 & $2.53 \times 10^{-6}$  & $1.27 \times 10^{-6}$  \\ \hline
[31,33]                    & 0.100418 & 0.102359 & $-5.05 \times 10^{-4}$ & $-2.53 \times 10^{-4}$ \\ \hline
[40,44]                    & 0.126007 & 0.128943 & $1.86 \times 10^{-2}$  & $4.65 \times 10^{-3}$  \\ \hline
[46,48]                    & 0.107663 & 0.100255 & $-1.78 \times 10^{-3}$ & $-8.90 \times 10^{-4}$ \\ \hline
[48,50]                    & 0.100255 & 0.104507 & $-1.01 \times 10^{-3}$ & $-5.06 \times 10^{-4}$ \\ \hline
[51,53]                    & 0.102917 & 0.105715 & $-5.80 \times 10^{-4}$ & $-2.90 \times 10^{-4}$ \\ \hline
[58,60]                    & 0.105192 & 0.093024 & $-5.87 \times 10^{-4}$ & $-2.93 \times 10^{-4}$ \\ \hline
[60,62]                    & 0.093024 & 0.100763 & $-1.68 \times 10^{-3}$ & $-8.39 \times 10^{-4}$ \\ \hline
[62,65]                    & 0.100763 & 0.104643 & $2.26 \times 10^{-3}$  & $7.53 \times 10^{-4}$  \\ \hline
[65,68]                    & 0.104643 & 0.108075 & $5.04 \times 10^{-4}$  & $1.68 \times 10^{-4}$  \\ \hline
[68,74]                    & 0.108075 & 0.091310 & $1.44 \times 10^{-2}$  & $2.40 \times 10^{-3}$  \\ \hline
[74,84]                    & 0.091310 & 0.059902 & $-4.97 \times 10^{-2}$ & $-4.97 \times 10^{-3}$ \\ \hline\hline
\end{tabular}
}
\caption{Numerical results of SurvSHAP for PC2 on the 3rd unit.}
\label{table_loc_300}
\end{minipage} \vfill
\vspace{5mm}
\begin{minipage}{0.68\textwidth}
\centering
\includegraphics[width=\linewidth]{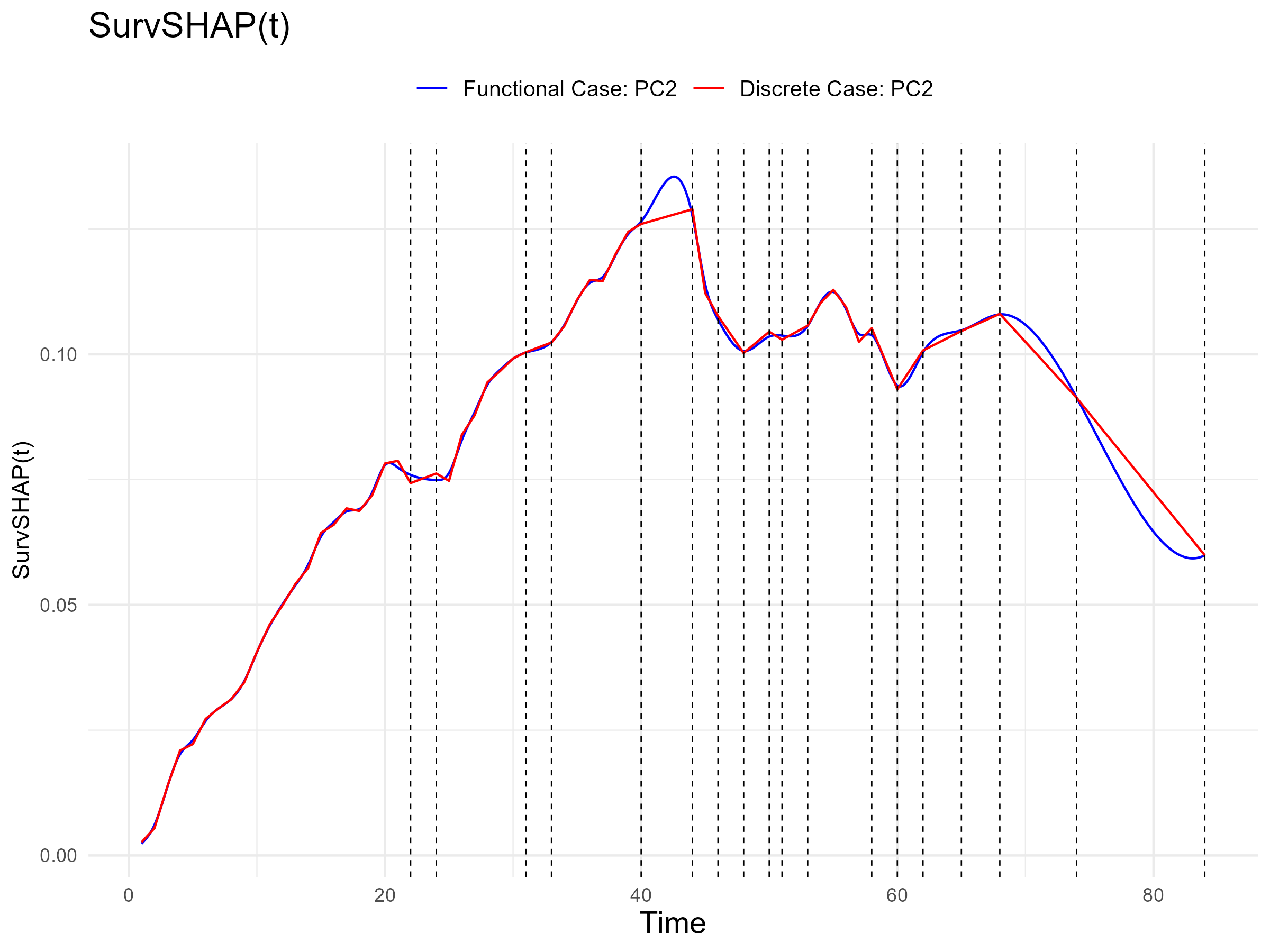}
\captionof{figure}{Local explainability of PC2 in FRSF for 3rd unit. The discrete case (red line) is compared to the functional reconstruction (blue line). Vertical black dashed lines indicate time intervals with no observed events.}
\label{fig:your_image4}
\end{minipage}
\end{table}

\clearpage

\subsection*{Application to SOFA Dataset}

\begin{table}
\Large
\centering
\begin{minipage}{0.70\textwidth}
\centering
\renewcommand{\arraystretch}{1.2} 
\setlength{\tabcolsep}{19pt} 
\resizebox{\textwidth}{!}{ 
\begin{tabular}{c|cc|c|c} 
\hline
\hline
\multicolumn{5}{c}{\textbf{Global Explainability for PC2}} \\  
\hline
\hline
\multirow{2}{*}{\([t_\alpha,t_\beta]\)} & \multicolumn{2}{c|}{\(\overline{FI}_2(t)\)} & \multirow{2}{*}{\(MTGD_{t_\alpha,t_\beta}^{2}\)} &\multirow{2}{*} {\(MTNGD_{t_\alpha,t_\beta}^{2}\)} \\ \cline{2-3}
                        & \(\overline{FI}_2(t_\alpha)\) & \(\overline{FI}_2(t_\beta)\) &  &  \\ 
\hline
[21,23]                 & 0.026781  & 0.026783  & $1.10 \times 10^{-4}$  & $5.48 \times 10^{-5}$  \\ \hline
[32,35]                 & 0.028364  & 0.028526  & $-6.48 \times 10^{-5}$ & $-2.16 \times 10^{-5}$ \\ \hline
[37,42]                 & 0.025973  & 0.024630  & $9.58 \times 10^{-4}$  & $1.92 \times 10^{-4}$  \\ \hline
[45,48]                 & 0.020825  & 0.019953  & $1.03 \times 10^{-4}$  & $3.42 \times 10^{-5}$  \\ \hline
[48,50]                 & 0.019953  & 0.018130  & $-1.08 \times 10^{-4}$ & $-5.39 \times 10^{-5}$ \\ \hline
[50,53]                 & 0.018130  & 0.018666  & $-5.53 \times 10^{-5}$ & $-1.84 \times 10^{-5}$ \\ \hline
[53,55]                 & 0.018666  & 0.018970  & $-2.37 \times 10^{-4}$ & $-1.19 \times 10^{-4}$ \\ \hline
[56,69]                 & 0.019051  & 0.019705  & $2.37 \times 10^{-2}$  & $1.82 \times 10^{-3}$  \\ \hline
[69,71]                 & 0.019705  & 0.014951  & $6.21 \times 10^{-5}$  & $3.10 \times 10^{-5}$  \\ \hline
[71,75]                 & 0.014951  & 0.013475  & $-4.48 \times 10^{-4}$ & $-1.12 \times 10^{-4}$ \\ \hline
[75,157]                & 0.013475  & 0.013671  & $1.19 \times 10^{-1}$  & $1.45 \times 10^{-3}$  \\ \hline
[157,173]               & 0.013671  & 0.005147  & $4.14 \times 10^{-3}$  & $2.59 \times 10^{-4}$  \\ \hline \hline
\end{tabular}
}
\caption{Numerical results in PFI for PC2.}
\label{table_glob_sofa}
\end{minipage} \vfill
\vspace{5mm}
\begin{minipage}{0.68\textwidth}
\centering
\includegraphics[width=\linewidth]{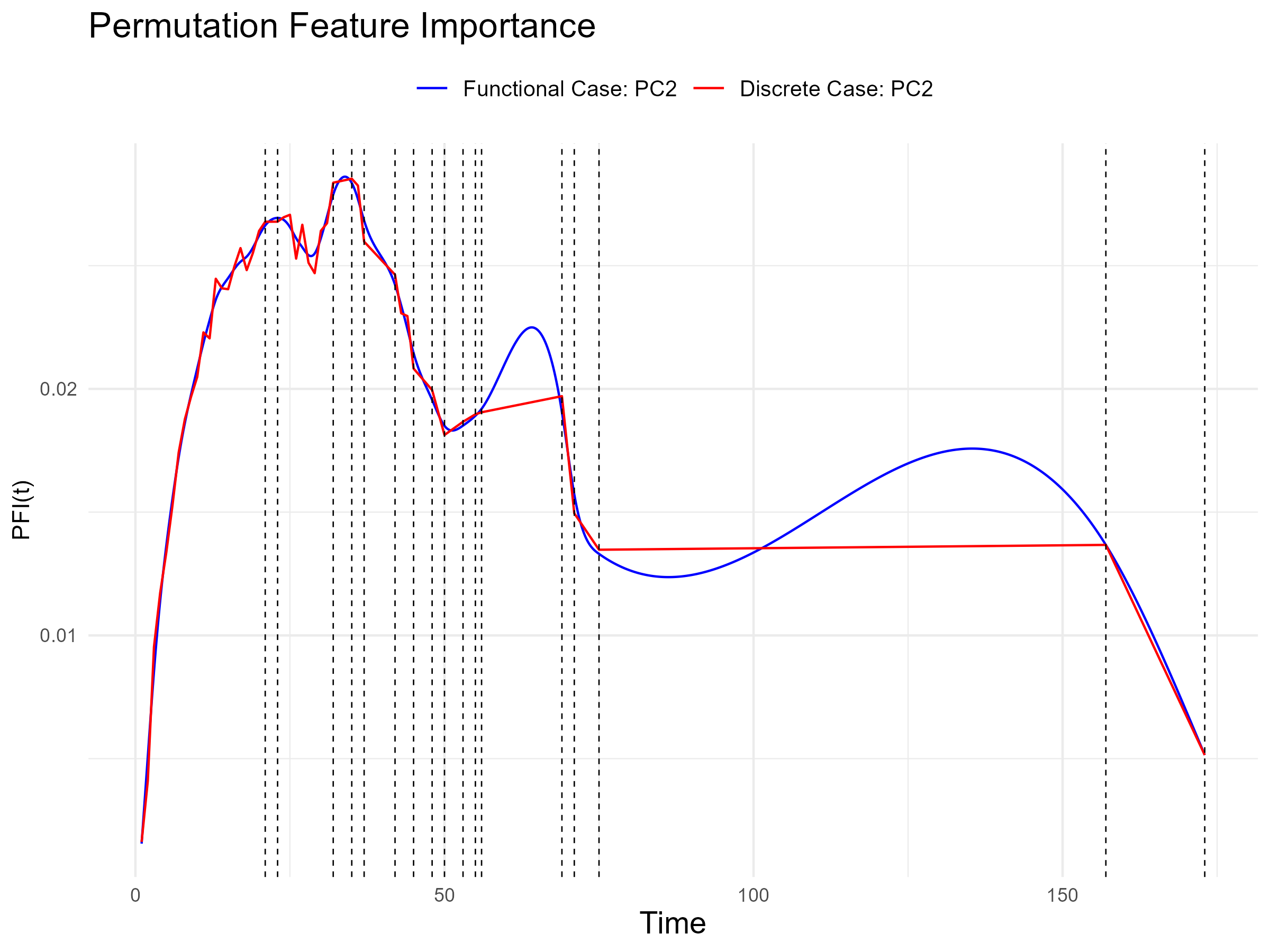}
\captionof{figure}{PFI-based global explainability of PC2 in FRSF, comparing the discrete case (red) and the functional reconstruction (blue). Black dashed lines mark intervals with no observed events.}
\label{recost_sofa_global}
\end{minipage}
\end{table}  

\begin{table}[h]
\Large
\centering
\begin{minipage}{0.70\textwidth}
\centering
\renewcommand{\arraystretch}{1.2} 
\setlength{\tabcolsep}{19pt} 
\resizebox{\textwidth}{!}{ 
\begin{tabular}{c|cc|c|c} 
\hline
\hline
\multicolumn{5}{c}{\textbf{Local Explainability for PC10 for $2$th unit}} \\  
\hline
\hline
\multirow{2}{*}{\([t_\alpha,t_\beta]\)} & \multicolumn{2}{c|}{\(\phi^*_t(\bm{w_*},10)\)} & \multirow{2}{*}{\(TSD_{t_\alpha,t_\beta}^{2\,10}\)} &\multirow{2}{*} {\(TNSD_{t_\alpha,t_\beta}^{2\,10}\)} \\ \cline{2-3}
                        & \(\phi^*_{t_\alpha}(\bm{w_*},10)\) & \(\phi^*_{t_\beta}(\bm{w_*},10)\) &  &  \\ 
\hline
[21,23]      & 0.0009116  & 0.0010929  & $6.24 \times 10^{-5}$  & $3.12 \times 10^{-5}$  \\ \hline
[32,35]      & -0.0003153 & -0.0002607 & $-3.66 \times 10^{-5}$ & $-1.22 \times 10^{-5}$ \\ \hline
[37,42]      & -0.0001365 & 0.0004267  & $-5.29 \times 10^{-4}$ & $-1.06 \times 10^{-4}$ \\ \hline
[45,48]      & 0.0014814  & 0.0017866  & $2.88 \times 10^{-4}$  & $9.60 \times 10^{-5}$  \\ \hline
[48,50]      & 0.0017866  & 0.0014227  & $-1.77 \times 10^{-6}$ & $-8.87 \times 10^{-7}$ \\ \hline
[50,53]      & 0.0014227  & 0.0016933  & $-1.21 \times 10^{-4}$ & $-4.02 \times 10^{-5}$ \\ \hline
[53,55]      & 0.0016933  & 0.0018889  & $5.20 \times 10^{-5}$  & $2.60 \times 10^{-5}$  \\ \hline
[56,69]      & 0.0014801  & 0.0011497  & $-9.24 \times 10^{-3}$ & $7.11 \times 10^{-4}$  \\ \hline
[69,71]      & 0.0011497  & 0.0018811  & $4.25 \times 10^{-5}$  & $2.12 \times 10^{-5}$  \\ \hline
[71,75]      & 0.0018811  & 0.0015163  & $5.31 \times 10^{-4}$  & $1.33 \times 10^{-4}$  \\ \hline
[75,157]     & 0.0015163  & 0.0024193  & $-1.22 \times 10^{-1}$ & $-1.49 \times 10^{-3}$ \\ \hline
[157,173]    & 0.0024193  & 0.0019311  & $1.02 \times 10^{-3}$  & $6.40 \times 10^{-5}$  \\ \hline\hline
\end{tabular}
}
\caption{Numerical results of SurvSHAP for PC10 on the 2nd unit.}
\label{table_loc_sofa}
\end{minipage} \vfill
\vspace{5mm}
\begin{minipage}{0.68\textwidth}
\centering
\includegraphics[width=\linewidth]{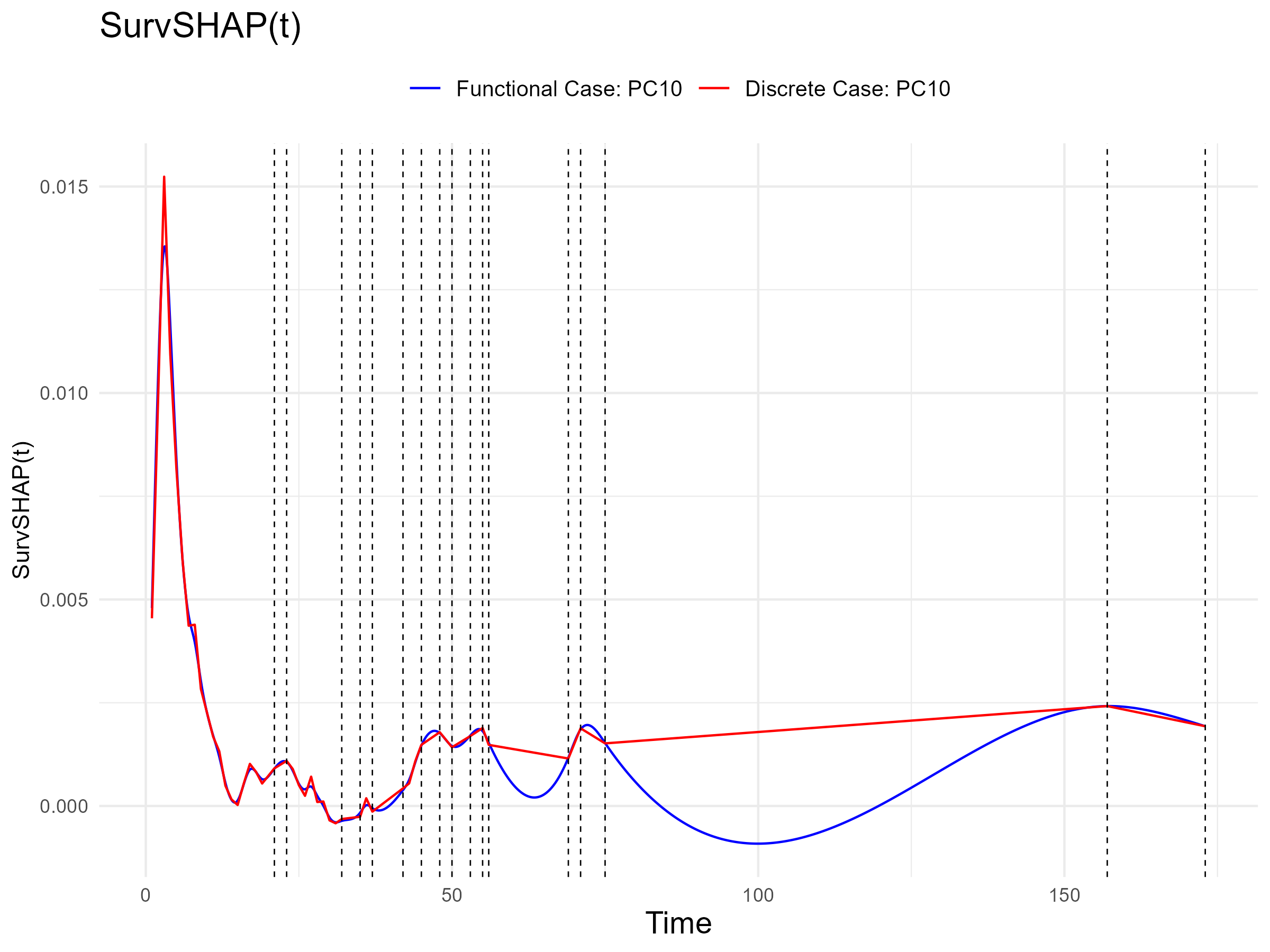}
\captionof{figure}{Local explainability of PC10 in FRSF for 2nd unit. The discrete case (red line) is compared to the functional reconstruction (blue line). Vertical black dashed lines indicate time intervals with no observed events.}
\label{recost_loc_sofa}
\end{minipage}
\end{table}


\end{document}